\documentclass[pdflatex,sn-mathphys]{sn-jnl}

\usepackage{subcaption}
\jyear{2022}%

\theoremstyle{thmstyleone}%
\theoremstyle{thmstyletwo}%

\theoremstyle{thmstylethree}%

\usepackage[utf8]{inputenc}
\usepackage[T1]{fontenc}
\usepackage{babel}
\usepackage{diagbox}
\usepackage[font=scriptsize]{caption} 
\usepackage{lineno}
\begin{document}

\title[Do Deep Neural Networks Always Perform Better When Eating More Data?]{Do Deep Neural Networks Always Perform Better When Eating More Data?}


\author[1]{\fnm{Jiachen} \sur{Yang}}
\author[1]{\fnm{Zhuo} \sur{Zhang}}
\author[1]{\fnm{Yicheng} \sur{Gong}}
\author[1]{\fnm{Shukun} \sur{Ma}}
\author[1]{\fnm{Xiaolan} \sur{Guo}}
\author[1]{\fnm{Yue} \sur{Yang}}
\author[1]{\fnm{Shuai} \sur{Xiao}}
\author[1]{\fnm{Jiabao} \sur{Wen}}
\author*[1,2]{\fnm{Yang} \sur{Li}}\email{liyang328@shzu.edu.cn}
\author[3]{\fnm{Xinbo} \sur{Gao}}
\author[3]{\fnm{Wen} \sur{Lu}}
\author[4]{\fnm{Qinggang} \sur{Meng}}

\affil[1]{\orgdiv{School of Electrical and Information Engineering}, \orgname{Tianjin University}, \orgaddress{\city{Tianjin}, \country{China}}}
\affil*[2]{\orgdiv{College of Mechanical and Electrical Engineering}, \orgname{Shihezi University}, \orgaddress{\city{Shihezi}, \country{China}}}
\affil[3]{\orgdiv{College of Mechanical and Electrical Engineering}, \orgname{Xidian University}, \orgaddress{\city{Xi'an}, \country{China}}}
\affil[4]{\orgdiv{College of Mechanical and Electrical Engineering}, \orgname{Loughborough University}, \orgaddress{\city{Loughborough}, \country{United Kingdom}}}


\abstract{Data has now become a shortcoming of deep learning. Researchers in their own fields share the thinking that "deep neural networks might not always perform better when they eat more data," which still lacks experimental validation and a convincing guiding theory. Here to fill this lack, we design experiments from Identically Independent Distribution(IID) and Out of Distribution(OOD), which give powerful answers. For the purpose of guidance, based on the discussion of results, two theories are proposed: under IID condition, the amount of information determines the effectivity of each sample, the contribution of samples and difference between classes determine the amount of sample information and the amount of class information; under OOD condition, the cross-domain degree of samples determine the contributions, and the bias-fitting caused by irrelevant elements is a significant factor of cross-domain. The above theories provide guidance from the perspective of data, which can promote a wide range of practical applications of artificial intelligence.}


\keywords{Data information, Independent and
identically distribution(IID), Out of distribution(OOD), Amount of information, Bias-fitting}

\maketitle

\section{Introduction}\label{sec1}
During the decade of the tremendous development of deep learning, it has made significance development in many fields\cite{lecun2015deep,webb2018deep,khoury2014big,lv2020solving,wang2019pseudo,li2020deep,yang2019fog,cao2022ai,yang2022survey,dong2021survey,qiu2021semantic}, such as agriculture, finance, medical health, and geography, which solves many complex pattern recognition problems\cite{esteva2021deep,aggarwal2021diagnostic,dai2021learning,zhao2018icnet,ren2017multi,pizzati2021comogan,bao2021evidential}. Current research and applications prefer to optimize the network to achieve higher accuracy instead of paying attention to the quality of datasets, though they are equally of great importance.\cite{shen2021towards,geirhos2020shortcut}. The neglect of attention to data leads to the following dilemmas which restrict the development and applications of deep learning.

Blindly expanding the amount of data not only increases the cost of hardware, environment, and resources, but also reduces the application efficiency in various fields\cite{gu2021pit,li2021semantic}. Deep learning requires train data with more contribution, and train data with little or no contribution will not only have little effect on the improvement of network performance, but also cause waste\cite{prakash2021multi, radford2019language, tan2021equalization}. If only increase the amount of data and ignore contribution, the network performance will not be satisfactory, and even result in high expenditure and low income. It can be seen that blindly pursuing the expansion of the amount of data does not always improve the performance of the network.

Inappropriate data can mislead the network, which effects performance and application. The deep learning network has a strong dependence on data\cite{chu2016data,zhou2021learning}, which does not entirely refer to quantity, but also contribution. Although the performance of deep learning network on large-amount public datasets is great\cite{van2021deep,yang2021mtd}, in some practical application scenarios\cite{ji2021learning,dourado2020open,tabernik2020segmentation,zhang2019weld,donato2021diverse,he2020different,fahimi2020generative}, the network can't achieve the expected effect due to limitation in data acquisition and insufficient coverage.

\begin{figure}[t!]
    \centering
    \begin{subfigure}[a]{0.92\textwidth}
           \centering
           \includegraphics[width=\textwidth]{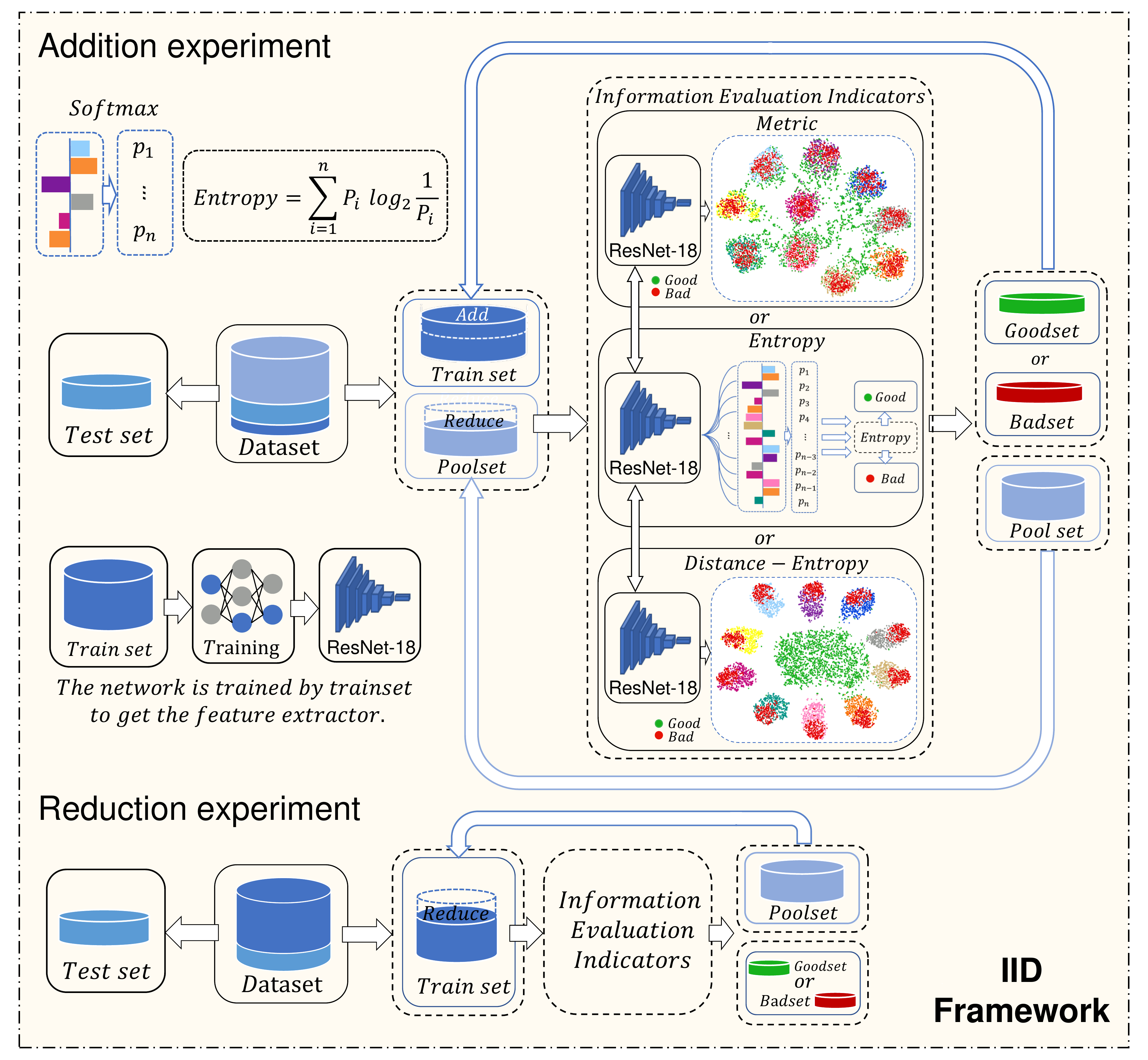}
            \caption{IID Framework}
            \label{IID-Framwork}
    \end{subfigure}
    \begin{subfigure}[b]{0.895\textwidth}
            \centering
            \includegraphics[width=\textwidth]{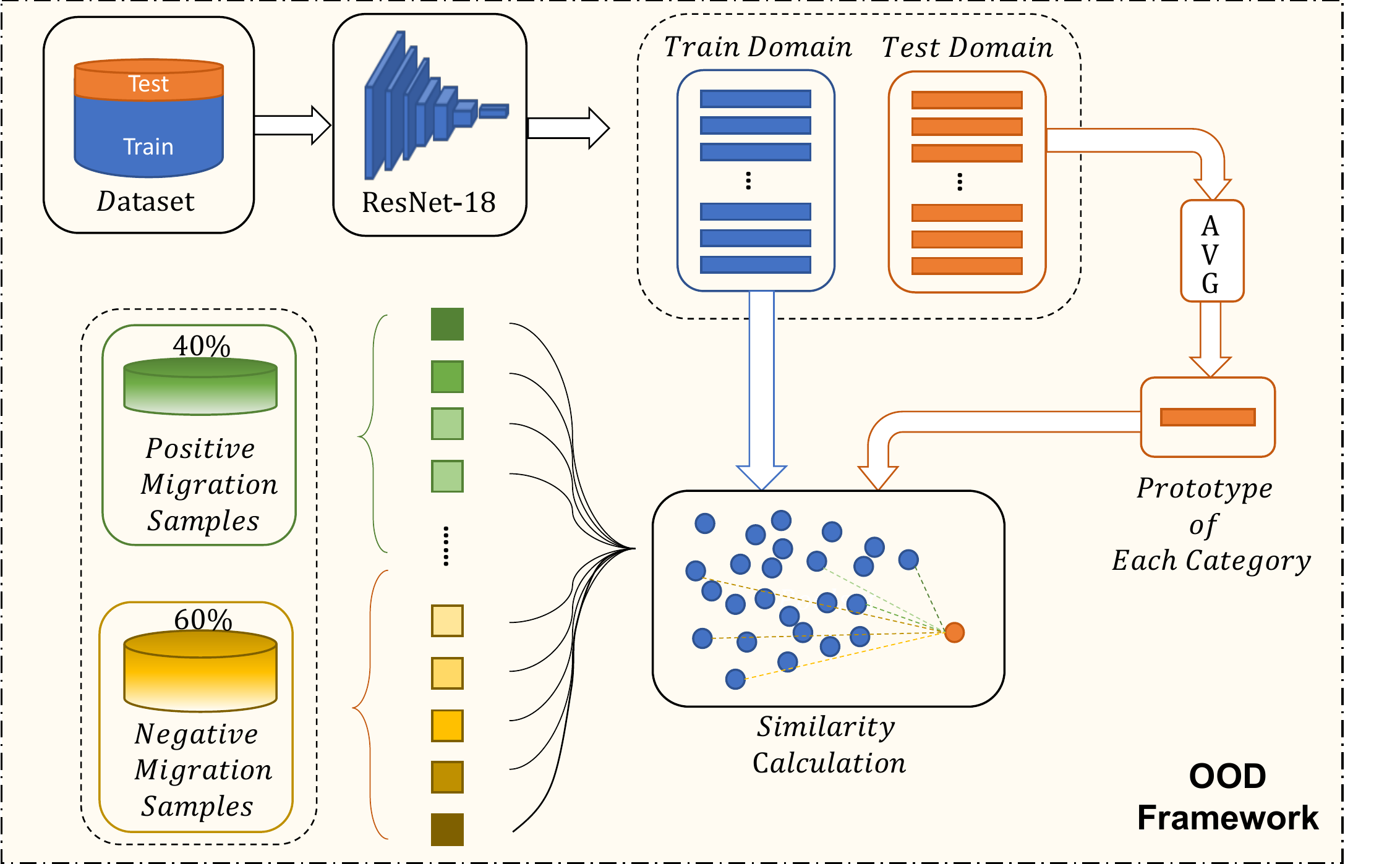}
            \caption{OOD Framework}
            \label{OOD-Framwork}
    \end{subfigure}
    \caption{The overall framework of the experiment.}
    \label{Framwork}
\end{figure}

Many researchers in various fields has begun to hold the opinion that "deep neural networks might not always perform better when they eat more data". However, with lack of experimental validation and  convincing guiding theories, the thought seldom attracts publicity. Firstly, if the extended train set and test set are independent and identically distribution (IID), the performance of the network actually contributed by the amount of information in train data. If the amount of information derived from sample is large enough, even if the number of data is small, a high-performance network can be trained. Secondly, if the extended train set and test set are out of distribution (OOD), network needs enough data closer to the test domain, which we call positive migration data, to achieve an ideal performance. More negative migration data even would reduce the overall performance of the network\cite{pan2009survey,huang2021fsdr}. We design several experiments in IID and OOD, the specific experimental overflow is shown in Fig. \ref{Framwork}.

In terms of IID, we select two common datasets, CIFAR-10 and Mini-ImageNet\cite{krizhevsky2009learning,vinyals2016matching}. There are three general Information Evaluation Indicators (IEIs) for assessing the amount of information: Distance Entropy\cite{li2021distance}, Entropy\cite{li2021entropy}, and Metric\cite{li2020few}. As shown in Fig. \ref{IID-Framwork}, these IEIs evaluate the sample contribution from the three aspects: network output distribution, inter-class relationship, and intra-class relationship. IEIs obtain and distinguish the contribution of each data. The experiment of gradually adding data (referred to "addition experiment") and reducing data (referred to "reduction experiment") are conducted, which proves that within an equal scale, a large amount of information data can achieve a better network training effect than a small amount of information data.

In terms of OOD, we select two datasets CSE and NICO-Animal\cite{he2021towards}, use Eculidean distance to distinguish the positive and negative migration data. By measuring the similarity of various feature vectors in the train domain and feature prototypes in the test domain\cite{snell2017prototypical}, we select data close to the test domain as positive migration data, and data far from the test domain as negative migration data. We make these two kinds of data into different two splits. The model performance of positive migration data split are better  than that of negative migration data split.

Based on these results, we further discuss and analyze the experimental phenomena in the fields of both IID and OOD. 

Efficient and feasible data selection currently lacks convincing guiding theories, perplexing researchers and engineers for a long time. In this paper, we
design experiments from IID and OOD, then provide a direction for problem above, it is that only a scientific data scheme can achieve an efficiency gain. In addition, we carry out discussion and verification, analyze the relationship between the amount of information in data and application, and discuss the causes of cross-domain phenomenon. The formed methodology can directly guide the work about data.

\section{Results}\label{sec2}

\subsection*{IID}\label{IID_result}
\begin{figure}[htbp]
	\centering
	\begin{subfigure}{0.38\linewidth}
		\centering
		\includegraphics[width=0.9\linewidth]{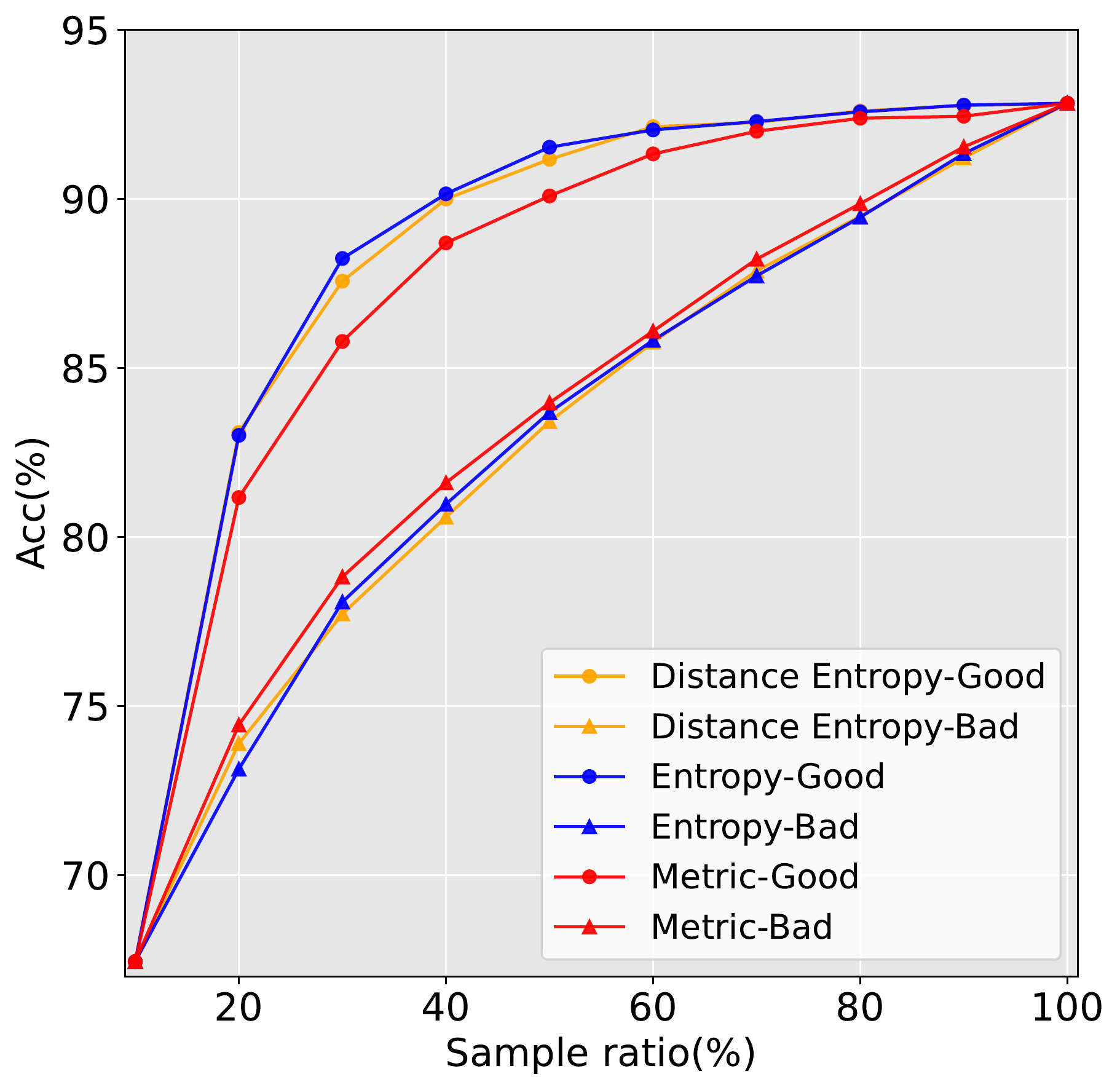}
		\caption{}
		\label{cifar10-addition}
	\end{subfigure}
	\centering
	\begin{subfigure}{0.38\linewidth}
		\centering
		\includegraphics[width=0.9\linewidth]{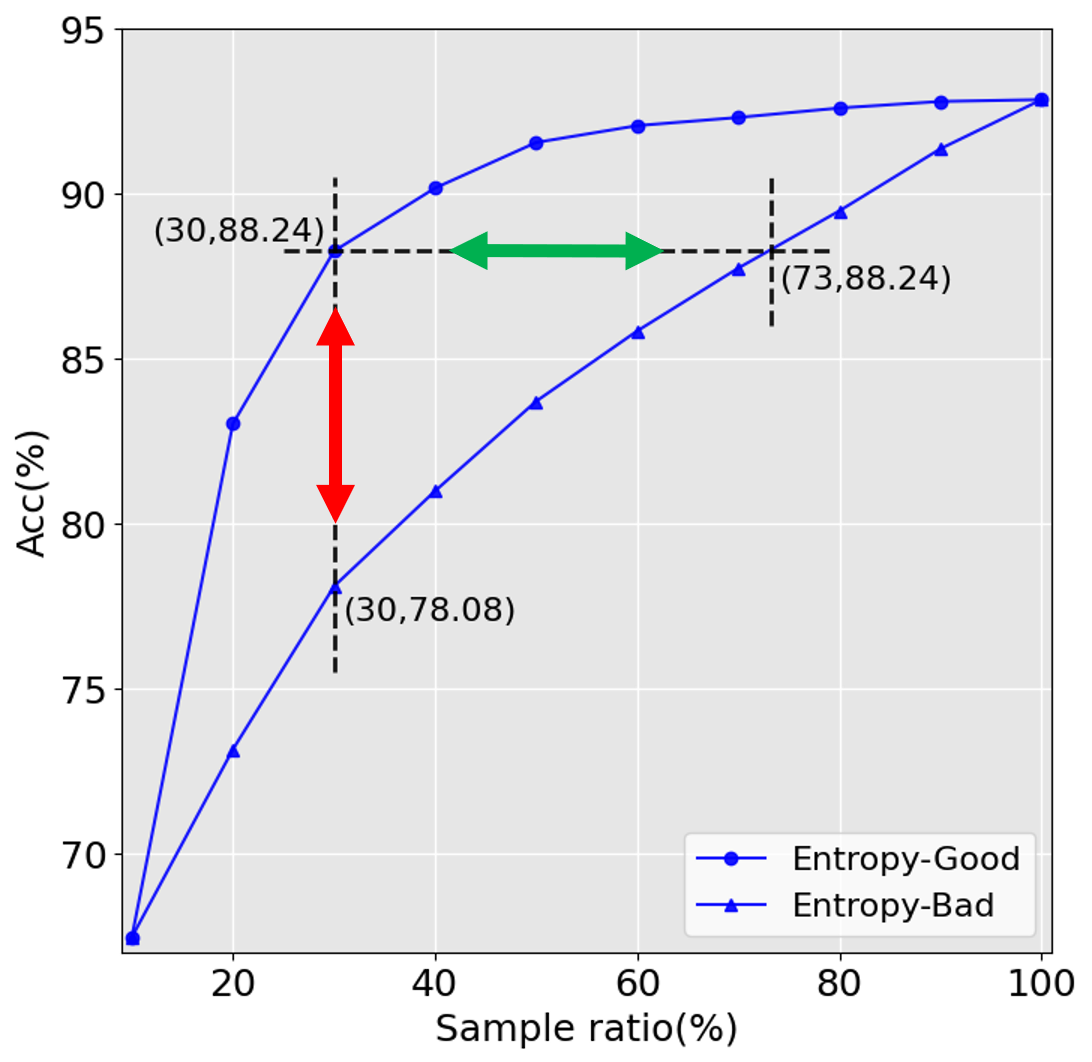}
		\caption{}
		\label{cifar10-addition-analysis}
	\end{subfigure}
	\centering
	\begin{subfigure}{0.38\linewidth}
		\centering
		\includegraphics[width=0.9\linewidth]{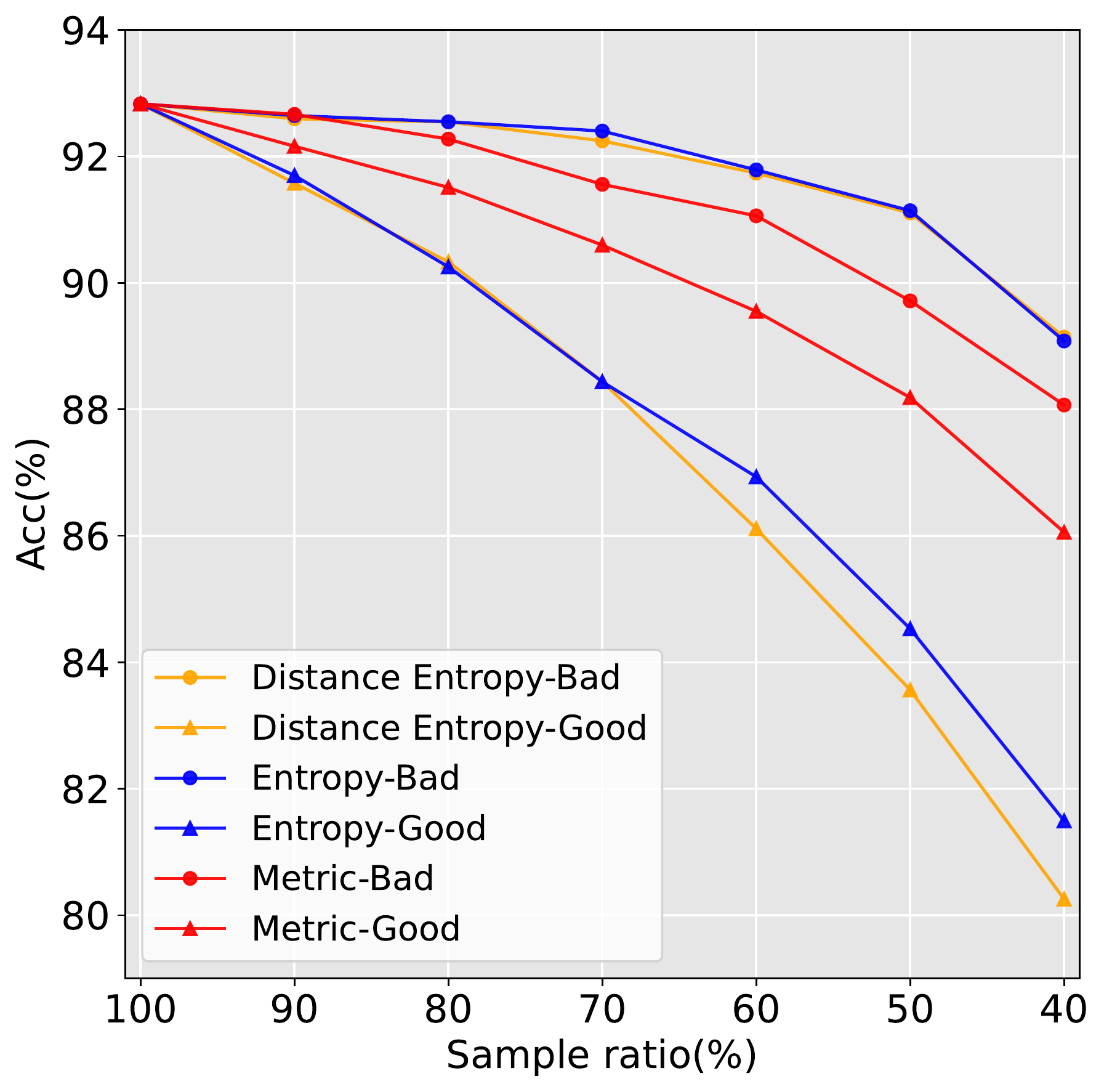}
		\caption{ }
		\label{cifar10-reduction}
	\end{subfigure}
	\begin{subfigure}{0.38\linewidth}
		\centering
		\includegraphics[width=0.9\linewidth]{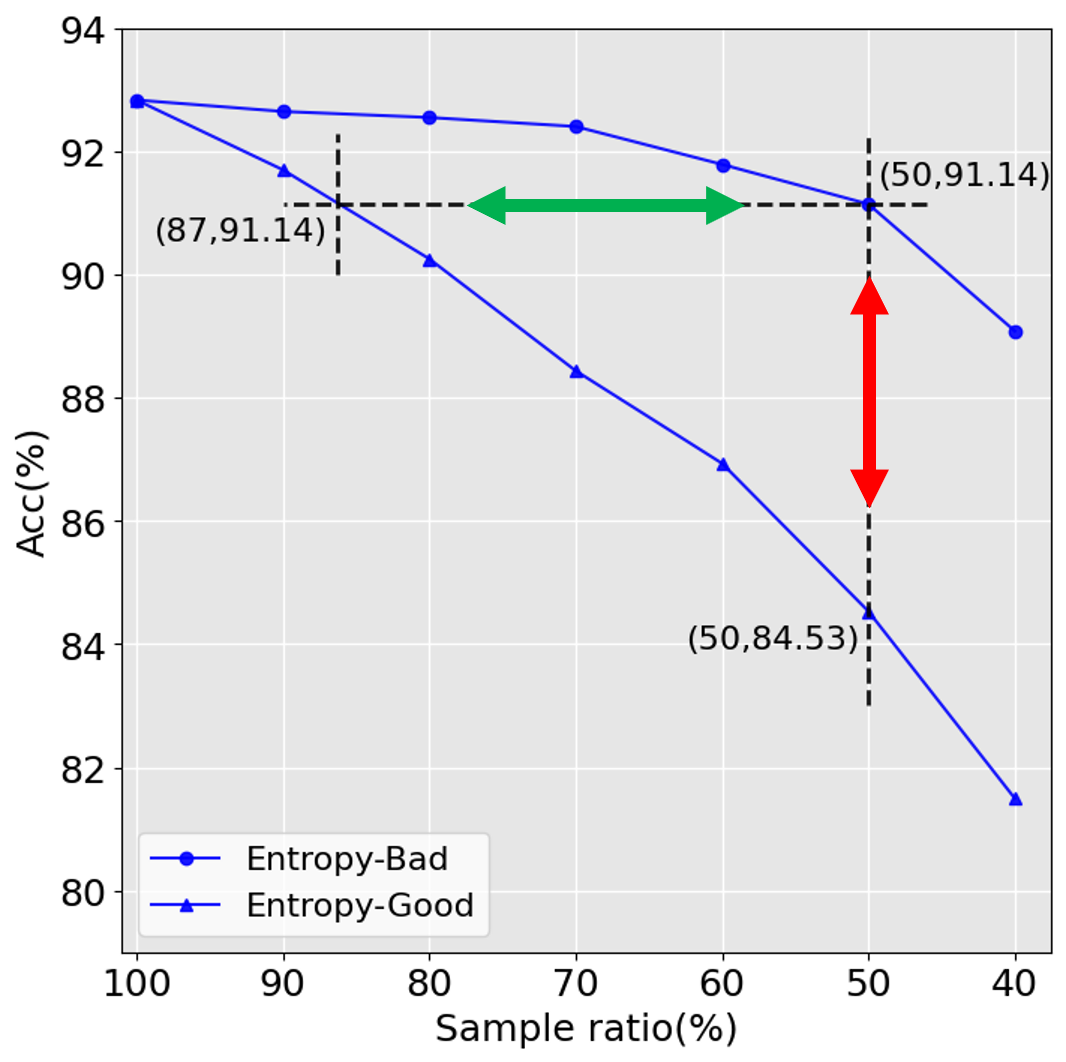}
		\caption{}
		\label{cifar10-reduction-analysis}
	\end{subfigure}

    	\centering
	\begin{subfigure}{0.38\linewidth}
		\centering
		\includegraphics[width=0.9\linewidth]{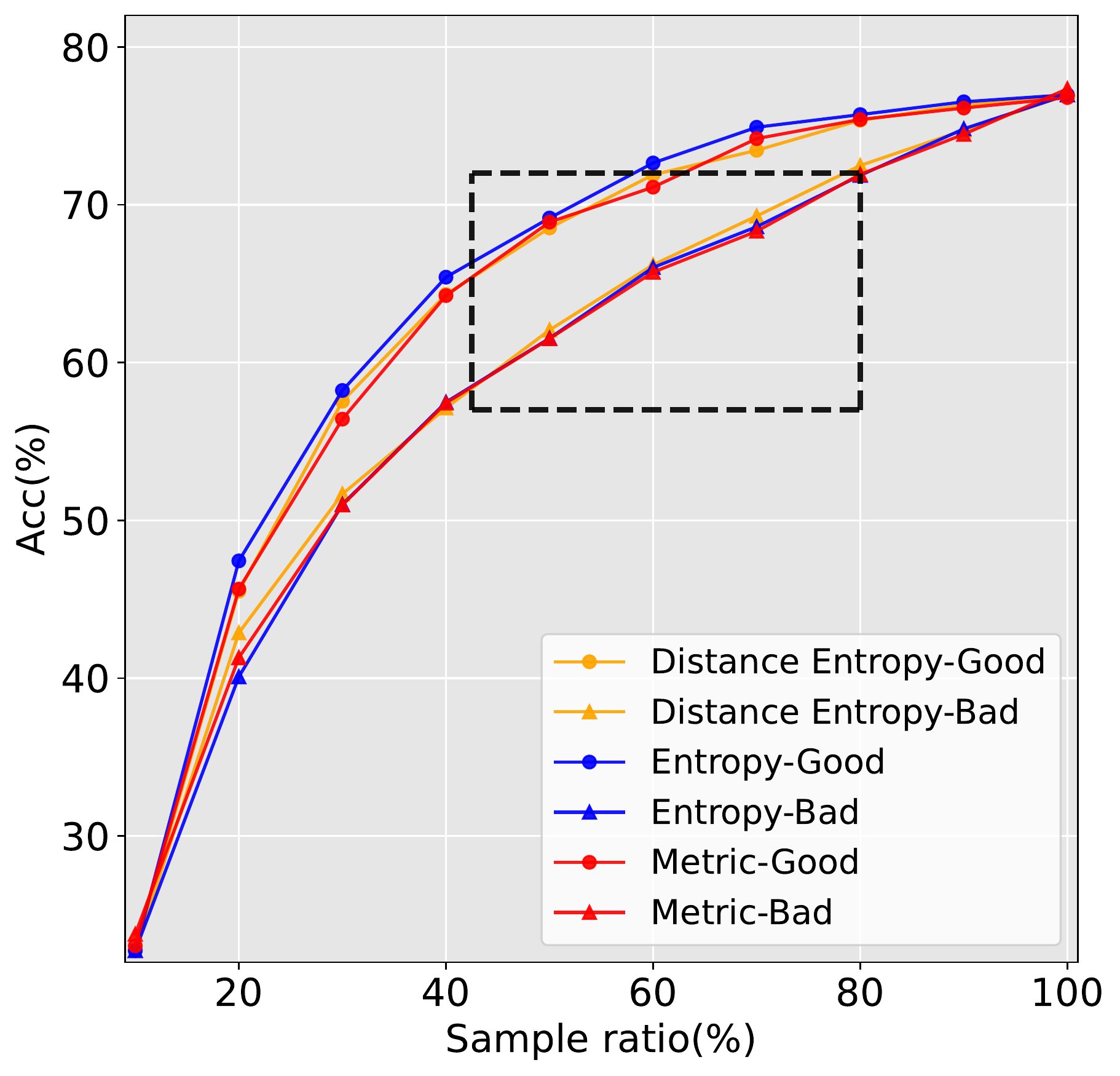}
		\caption{}
		\label{mini-addition}
	\end{subfigure}
	\centering
	\begin{subfigure}{0.38\linewidth}
		\centering
		\includegraphics[width=0.9\linewidth]{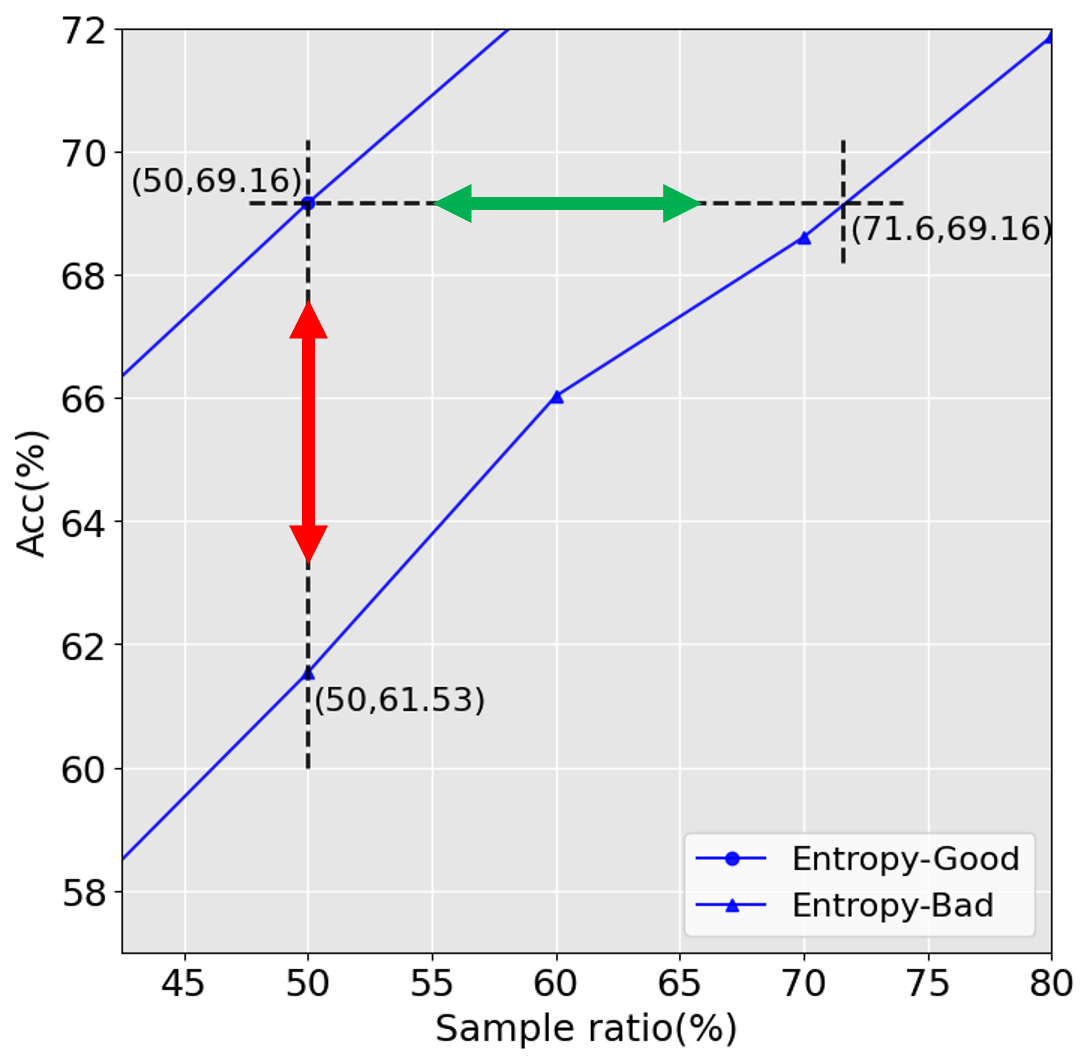}
		\caption{}
		\label{mini-addition-analysis}
	\end{subfigure}
	\centering
	\begin{subfigure}{0.38\linewidth}
		\centering
		\includegraphics[width=0.9\linewidth]{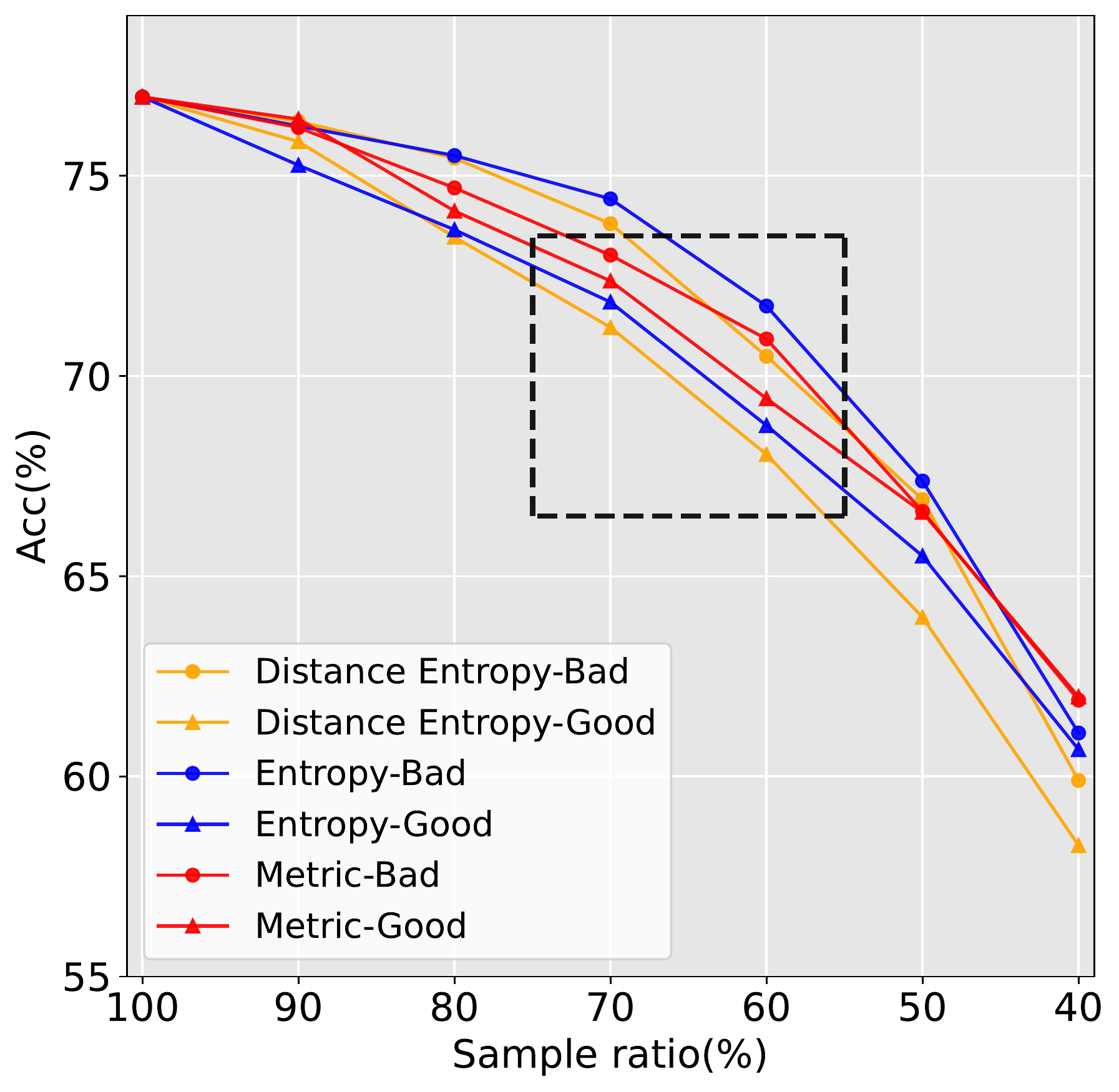}
		\caption{ }
		\label{mini-reduction}
	\end{subfigure}
	\begin{subfigure}{0.38\linewidth}
		\centering
		\includegraphics[width=0.9\linewidth]{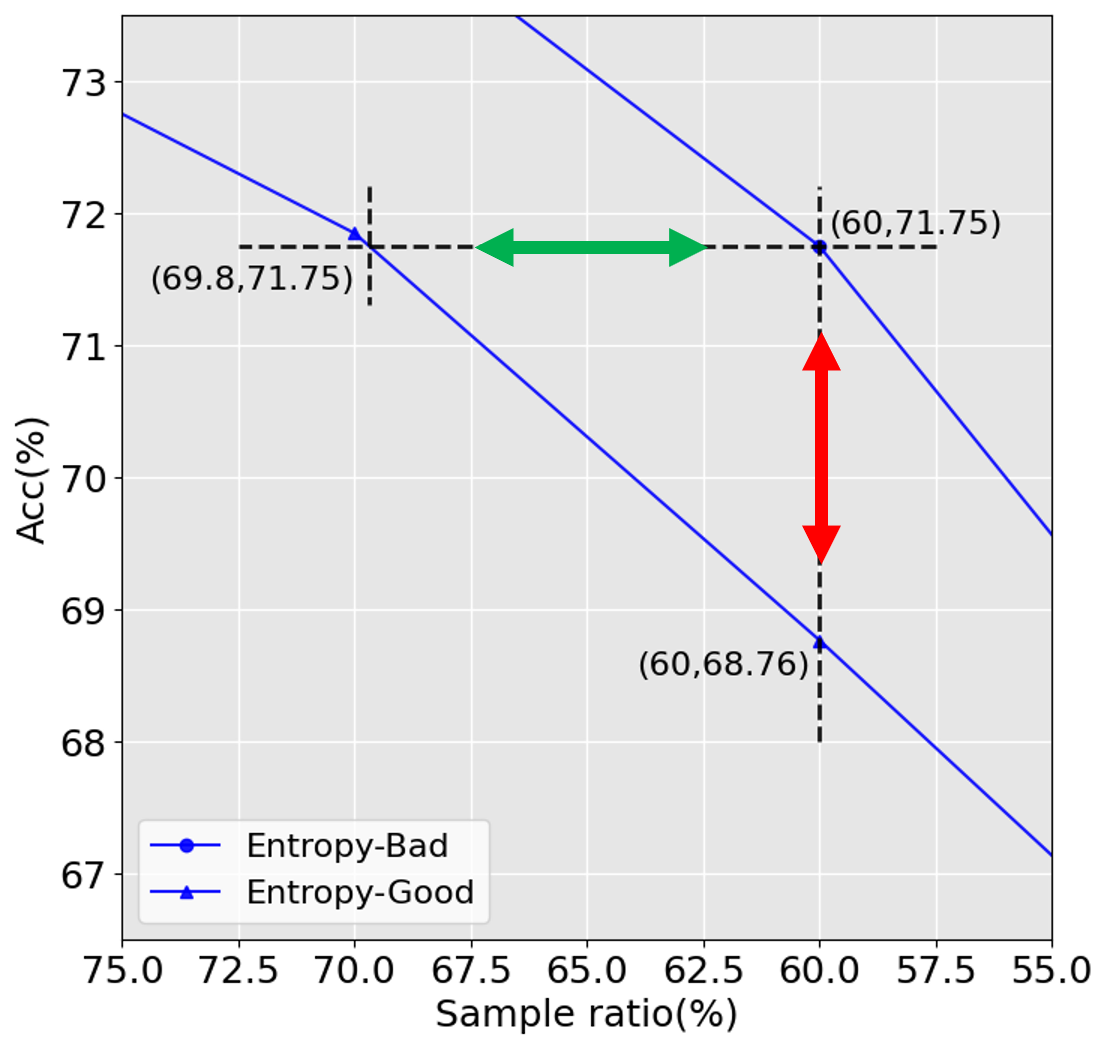}
		\caption{}
		\label{mini-reduction-analysis}
	\end{subfigure}
	\caption{The IID experiment results. (a)Results of "addition experiment" by different IEIs on CIFAR-10. (b)Analysis of the results of the "addition experiment" on CIFAR-10. (c)Results of "reduction experiment" by different IEIs on CIFAR-10. (d)Analysis of the results of the "reduction experiment" on CIFAR-10. (e)Results of "addition experiment" by different IEIs on Mini-ImageNet. (f)Analysis of the results of the "addition experiment" on Mini-ImageNet. (g)Results of "reduction experiment" by different IEIs on Mini-ImageNet. (h)Analysis of the results of the "reduction experiment" on Mini-ImageNet. }
	\label{IID Experiment Result}
\end{figure}

The purpose of the "addition experiment" and "reduction experiment" is to compare the performances of datasets with different information densities in the task by dynamically changing the amount of data. This experiment is based on image classification datasets CIFAR-10 and Mini-ImageNet and used ResNet-18 as the backbone network, which is widely accepted in the science and industry. To verify from multiple aspects, we use the IEIs to distinguish the data, and take samples with large amount of information as goodset, and vice versa as badset. The experimental results can  demonstrate the difference in network performance caused by different amount of information data. Fig. \ref{IID Experiment Result} shows the final results of two experiments under different IEIs, using CIFAR-10 and Mini-ImageNet. To reduce the experimental error, the experimental results are the average value of three repeated experiments.

\noindent\textbf{Addition experiment.}
Firstly, we take 10\% of the train set as the base for preliminary training, the rest would be the pool set to wait for selection. And the trained network is used as the feature extractor to evaluate the  amount of information derived from samples in the pool set. Based on the principle of selecting the samples with the larger amount of information, we select a goodset with a fixed budget and an even number of classes. The goodset would be added to the train set for the next training, which could build a dataset with high information density (HID dataset). Thus, the HID dataset would be added several times according to the above scheme, and each time the network would be retrained, the final result under different data budgets would be obtained. Selecting badset to build a dataset with low information density (LID dataset) and obtain the result could realize in an inverse selection principle. In order to verify the generalizability of the results on different IEIs, by changing the IEI, multiple experiments can be performed and the results obtained. Fig. \ref{cifar10-addition} shows the rise of the accuracy of HID dataset and LID datasets acquired by different IEIs on CIFAR-10. It can be seen that the performance of HID dataset has been better improved under different selection methods. As shown in Fig. \ref{cifar10-addition-analysis}, take the Entropy IEI as an example, when the network accuracy of the HID and LID dataset is 88.24\%, the LID dataset requires about 21500 additional samples. When the Sample ratio is 30\%, the network accuracy of the HID is 10.16\% higher than that of the LID dataset. 

\noindent\textbf{Reduction experiment.}
Firstly, we take all the data in the train set as the base for preliminary training. And the trained network is used as the feature extractor to evaluate the amount of information derived from sample in the train set. Based on the principle of selecting the samples with the smaller amount of information, we select a badset with a fixed budget and an even number of classes. The badset would be removed from the train set for the next training, which could achieve an HID datasets. Thus, the HID datasets would be reduced several times according to above scheme, and in each time the network would be retrained, the finally result under different data budgets would be obtained. Achieving a LID dataset and obtaining the results could be realized in an inverse reduction principle. Replace the IEIs, multiple experiments and results could be conducted and obtained. Fig. \ref{cifar10-reduction} shows the decline of Acc of goodset and badset reduced by different IEIs on CIFAR-10. It can be seen that under different IID IEIs, the performance of HID datasets decreased very slowly, while the performance of the LID dataset decreased sharply. As shown in  Fig. \ref{cifar10-reduction-analysis},  take the Entropy IEI as an example, when the network accuracy of the LID and HID dataset is 91.14\%, the HID dataset has about 18500 fewer samples than the LID dataset. When the Sample ratio is 50\%, the network accuracy of the HID is 6.61\% higher than that of the LID dataset. Fig. \ref{mini-addition}, Fig. \ref{mini-addition-analysis}, Fig. \ref{mini-reduction}, and Fig. \ref{mini-reduction-analysis} show the experimental results on Mini-ImageNet, which are the same as those on CIFAR-10.

In summary, through "addition experiment" and "reduction experiment" , we have come to the following conclusions: the samples are different in the amount of information, a smaller HID dataset can achieve the same effect as a larger LID dataset. Therefore, when in IID condition, due to the influence of the amount of information, the neural network may not achieve better performance by eating more data.

\subsection*{OOD}\label{OOD_result}
\begin{figure}[htbp]
	\centering
	\begin{subfigure}{0.55\linewidth}
		\centering
		\includegraphics[width=0.99\linewidth]{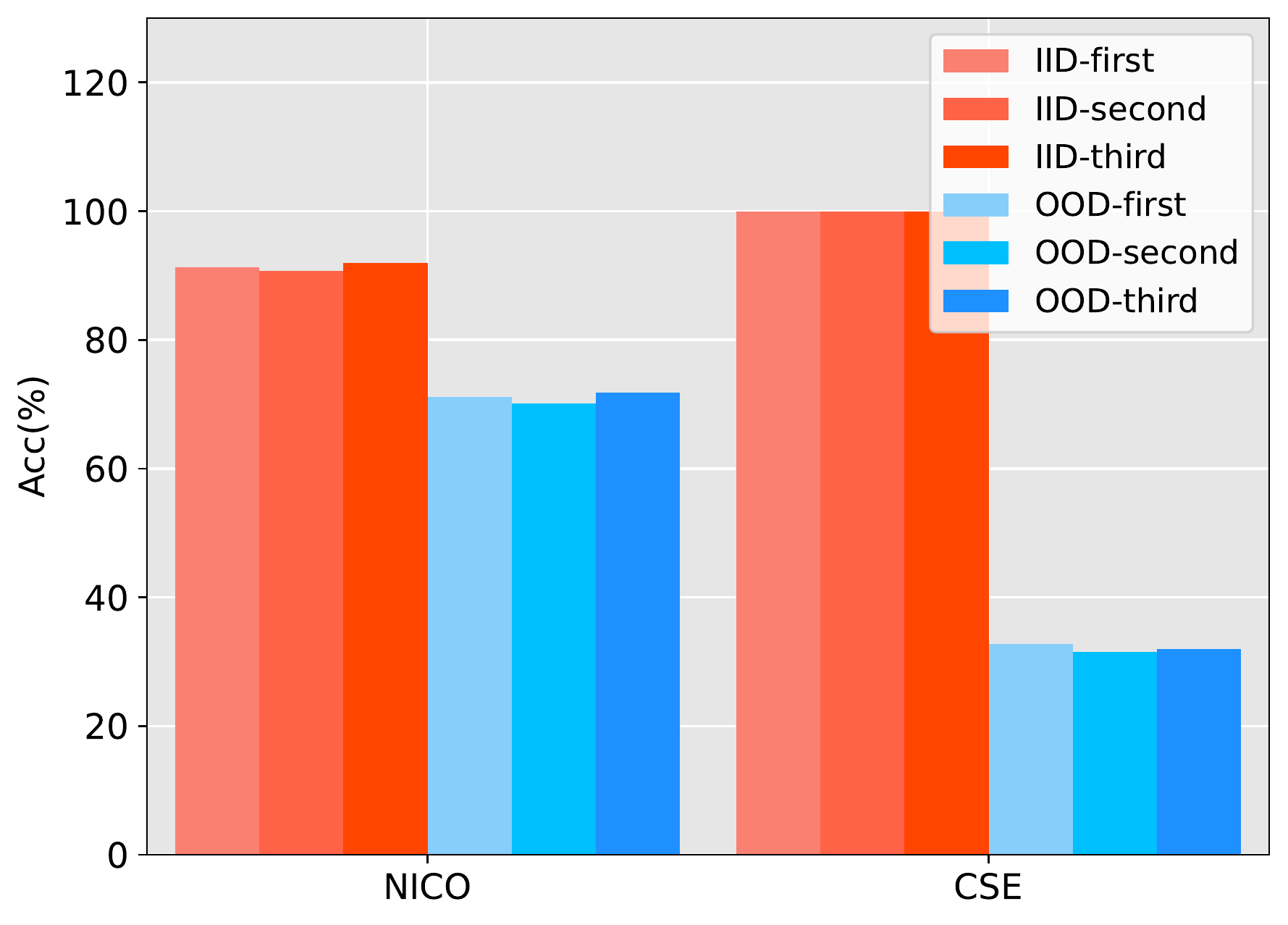}
		\caption{}
		\label{IID_OOD}
	\end{subfigure}
	\centering
	\begin{subfigure}{0.48\linewidth}
		\centering
		\includegraphics[width=0.99\linewidth]{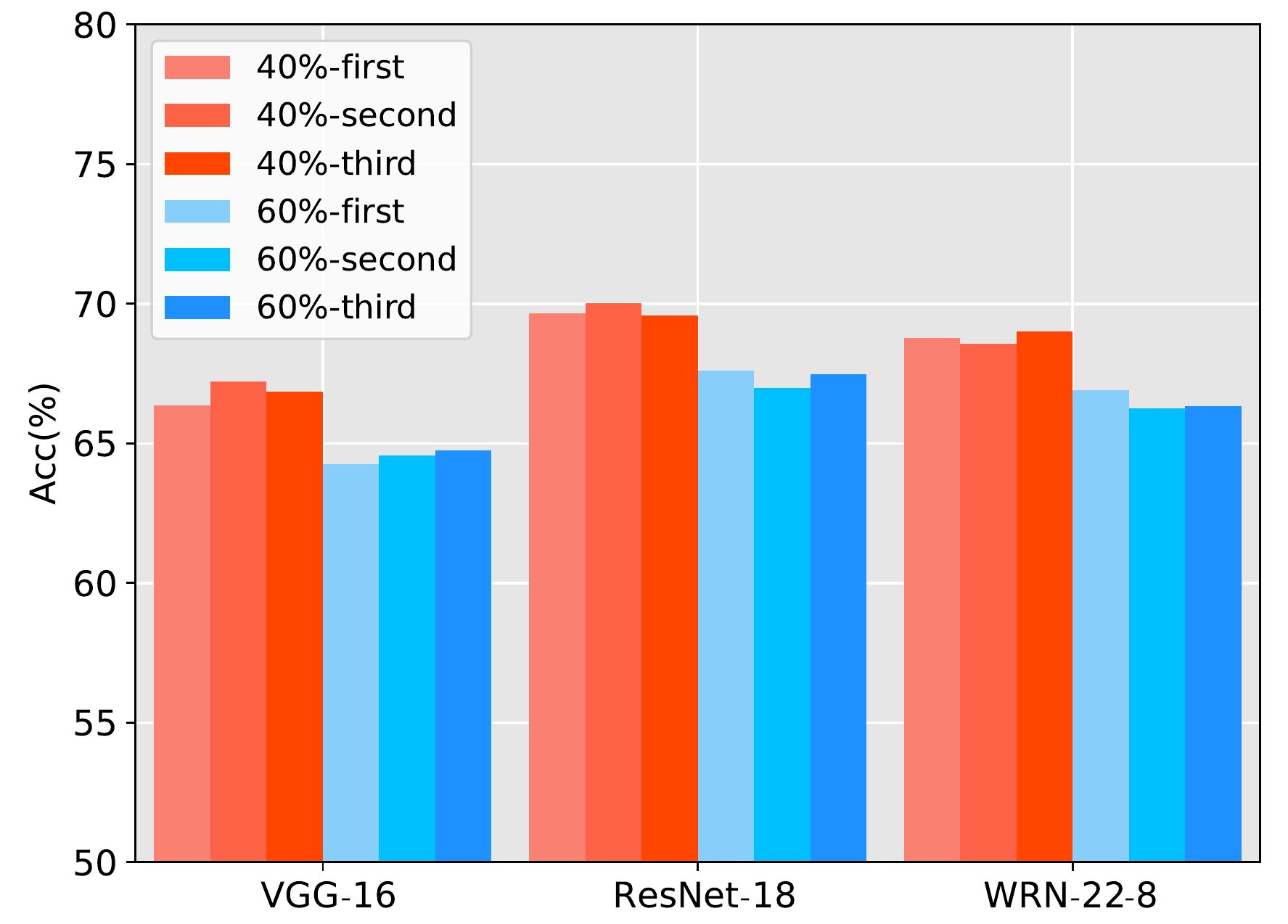}
		\caption{}
		\label{NICOOOD}
		
	\end{subfigure}
	\centering
	\begin{subfigure}{0.48\linewidth}
		\centering
		\includegraphics[width=0.99\linewidth]{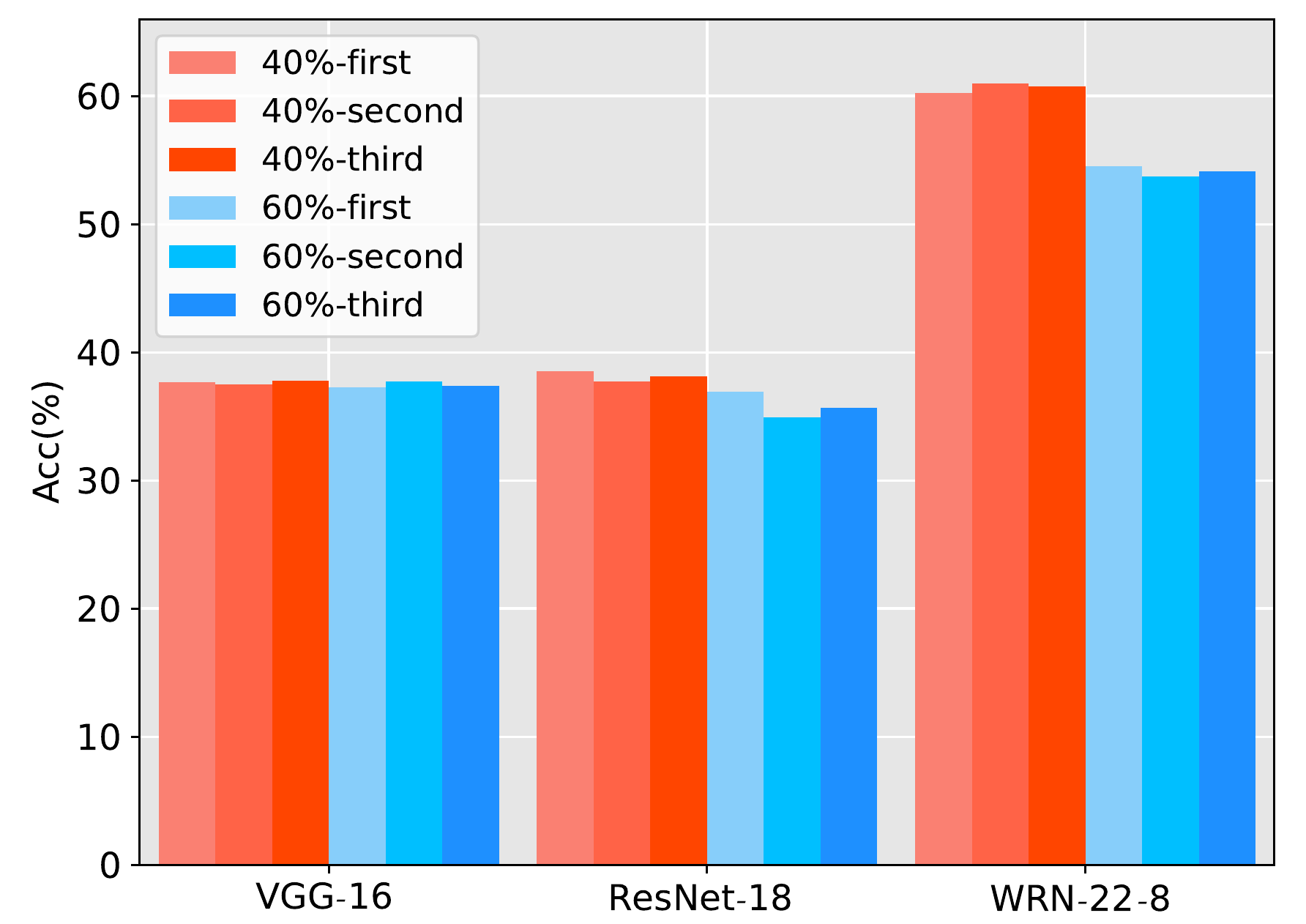}
		\caption{}
		\label{CSEOOD}
	\end{subfigure}
	\caption{The OOD experiment results. (a)Performance comparison between IID and OOD. (b)Results of three repeated experiments on NICO-Animal dataset. (c)Results of three repeated experiments on CSE dataset.}
	\label{OOD_Experiment}
\end{figure}

The purpose of the cross-domain split experiment is to compare the performance of the datasets with different cross-domain degrees. To verify the generalization of our method, we use three backbones of different sizes (ResNet-18, VGG-16, WRN-22-8) to get the performance of two cross-domain datasets (NICO-Animal, CSE). To ensure the scientificity of the experiments, we first verify the cross-domain phenomenon in the datasets. Then, the 
train set is splited by measuring the similarity between the feature vector of train domain's data and the feature prototype of test domain's data. We take 40\% data closer to the test domain as the positive migration data, and the remaining 60\% as the negative migration data. The experimental results in Fig. \ref{OOD_Experiment} show the difference of network performance caused by different cross-domain datasets.

\noindent\textbf{Cross-domain phenomenon of datasets.}
Before carrying out OOD experiments, we first verify the cross-domain attributes of the two datasets. Specifically, in NICO-Animal, we randomly select three contexts from each category to form the test set, and the remaining seven contexts of each class to form the trainval set. Then we randomly select 90\% of trainval set as the train set, and the remaining 10\% as the validation set. In CSE, 90\% of each class is selected as the train set and the other 10\% as the validation set. In the two datasets, test set is OOD with train set and validation set, and train set and validation set are IID with each other. To avoid the randomness of the experimental results, we train each group by ResNet-18 three times and the results are shown in Fig. \ref{IID_OOD}. It can be seen that when the dataset is IID distributed, the accuracy can reach more than 90\% in NICO-Animal and 100\% in CSE. When the dataset is OOD distributed, the accuracy is about 71\% in NICO-Animal and only about 32\% in CSE.

\noindent\textbf{Cross-domain split experiments.}
In two cross-domain datasets, we use ResNet-18 to obtain a feature extractor to extract the feature vector of the data in the trainval set and the feature prototype of the data in the test set. Calculate the similarity between all train domain samples and their corresponding class feature prototypes. After sorting the similarity from large to small, take the first 40\% of the data as the positive migration samples and the last 60\% as the negative migration samples. The two splits are respectively trained by three different backbones. Same as the previous experiment, we train each split three times to avoid the randomness, and the results are shown in Fig. \ref{NICOOOD}, \ref{CSEOOD}. It can be seen that in the experimental results under the three networks, except when training CSE with VGG-16 and ResNet-18, the accuracy of 40\% positive migration samples is 0.2\%$\sim$1\% higher than that of 60\% negative migration samples. In other experiments, the accuracy of the 40\% split can be more than 2\% higher than that of 60\% split. 

Through these cross-domain split experiments, the influence of cross-domain phenomenon and cross-domain degree on the performance of the network is proved. Under the condition of OOD, there is a fact that the performance does not always get better when the network is fed more samples.

\section{Discussion}\label{sec3}

For a long time, increasing samples to improve the performance of tasks is a common scheme. This also makes the recent deep learning researches gradually develop in the direction of big data and big networks\cite{sun2017revisiting,bommasani2021opportunities,korngiebel2021considering}. However, our research provides a new perspective for exploring data-driven algorithms: (i) under the condition of IID, the amount of information requires R\&D personnel to build datasets purposefully rather than supplement blindly, (ii) under the condition of OOD, pay attention to the cross-domain data in the field of data-driven algorithms could avoid introducing a large number of OOD data, resulting in the weakening of the algorithms.

The experimental results and analysis in section Result\ref{sec2} also demonstrate the correctness of our two assumptions, which can support our answer. However, the specific reasons for these experimental phenomenona still need to be discussed.

\subsection*{Two-tier amount of information theory under IID condition}

The phenomenon of the experiment in section \ref{IID_result} IID illustrates that amount of information derived from samples play a significant role to determine how much performance the network would improve. The more abundant the amount of information on the train set, the better the performance. However, how to embody the concept of information in the data is still an unanswered question.

\begin{figure}[htbp]
	\begin{subfigure}{0.48\linewidth}
		\centering
		\includegraphics[width=0.95\linewidth]{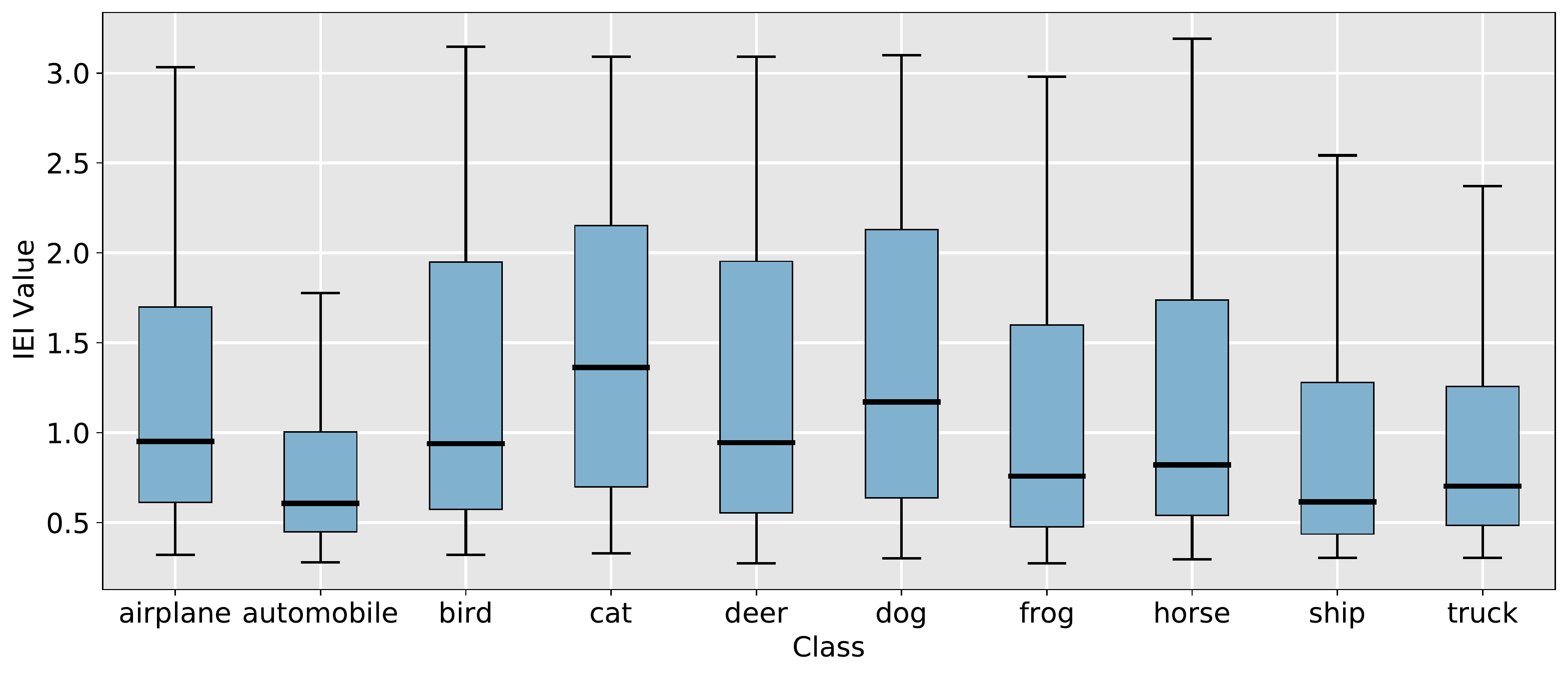}
		\caption{}
		\label{Figure4a}
	\end{subfigure}
	\centering
	\begin{subfigure}{0.48\linewidth}
		\centering
		\includegraphics[width=0.95\linewidth]{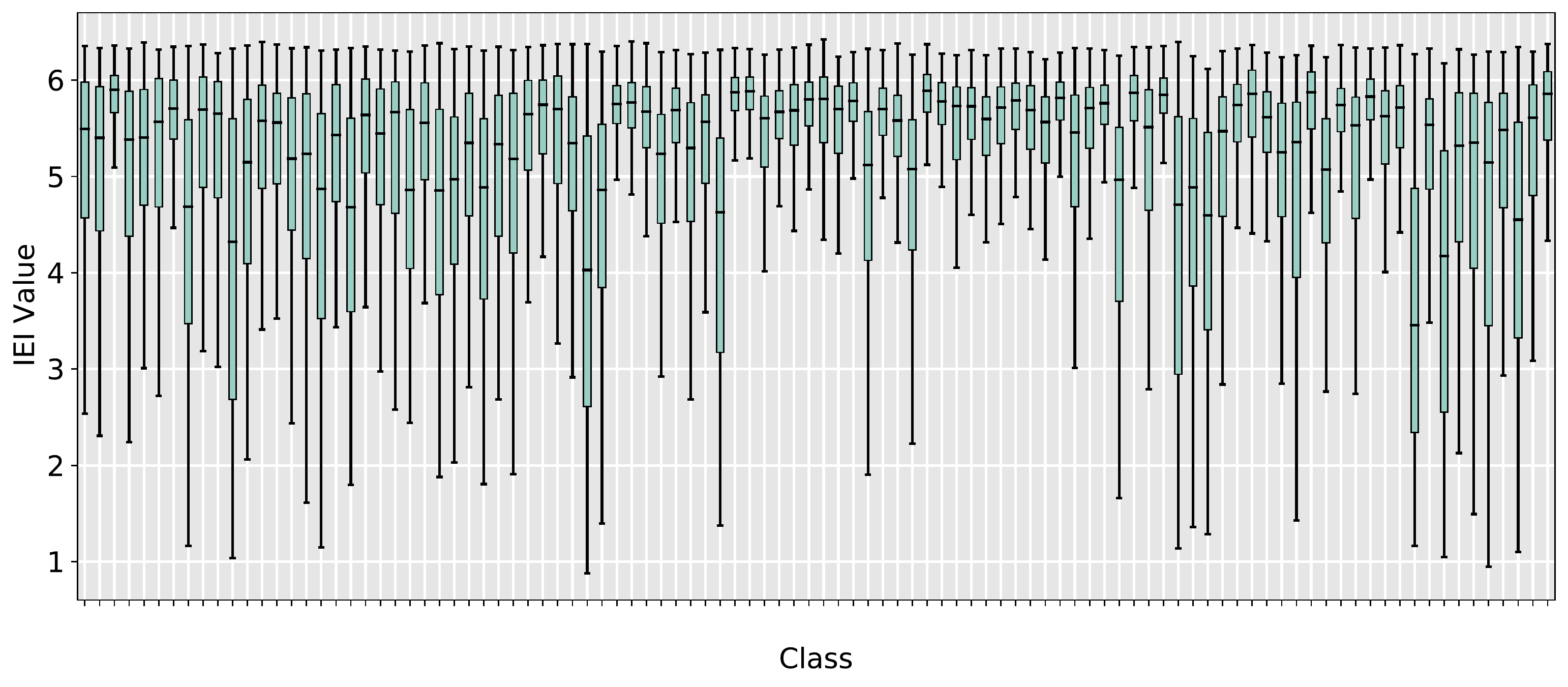}
		\caption{}
		\label{Figure4b}
	\end{subfigure}
	\centering
	\begin{subfigure}{0.48\linewidth}
		\flushleft
		\includegraphics[width=1\linewidth]{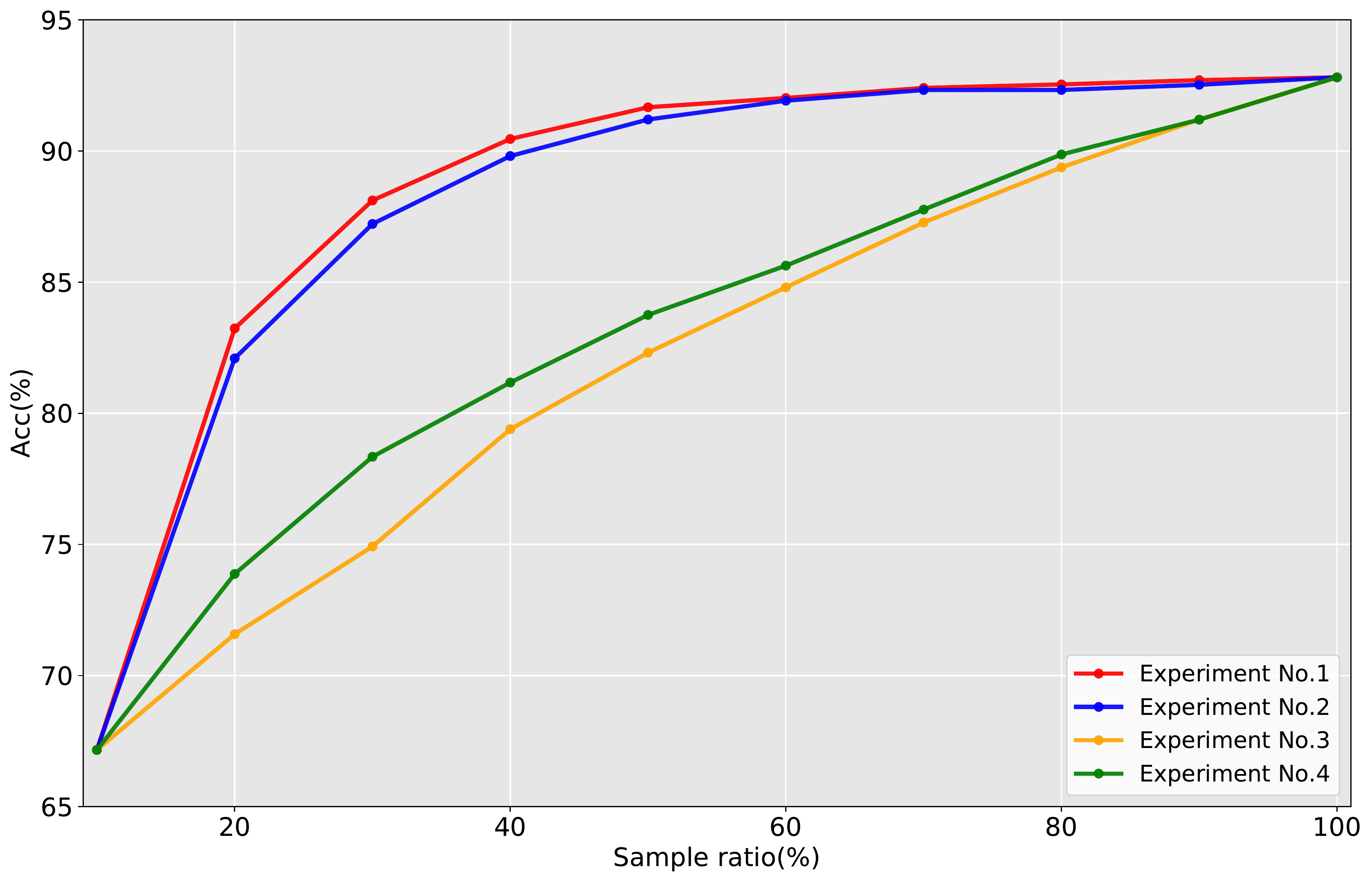}
		\caption{}
		\label{Figure4c}
	\end{subfigure}
	\centering
	\begin{subfigure}{0.48\linewidth}
		\flushright
		\includegraphics[width=1\linewidth]{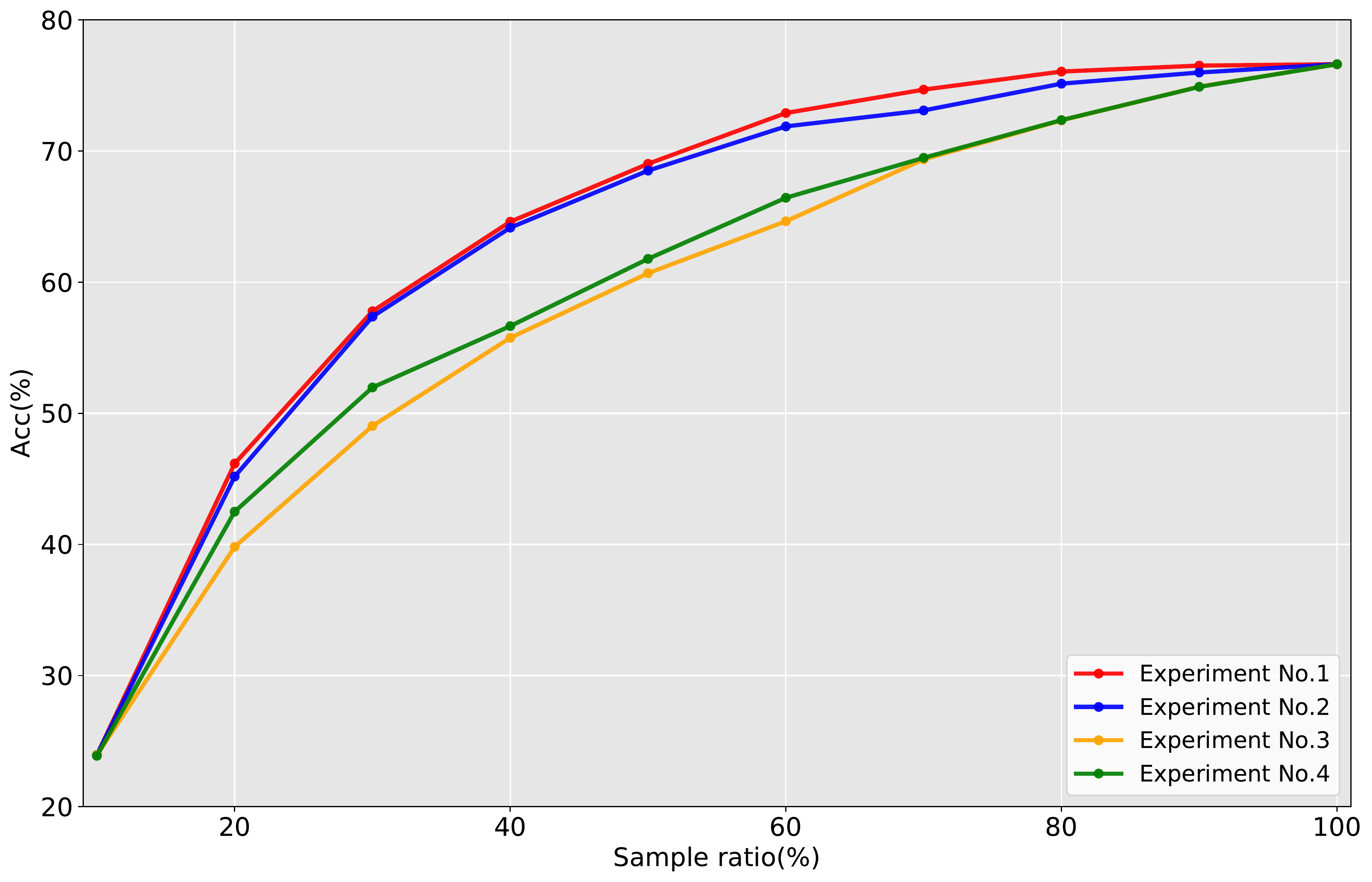}
		\caption{}
		\label{Figure4d}
	\end{subfigure}
	\centering
	\begin{subfigure}{0.96\linewidth}
		\centering
		\includegraphics[width=0.95\linewidth]{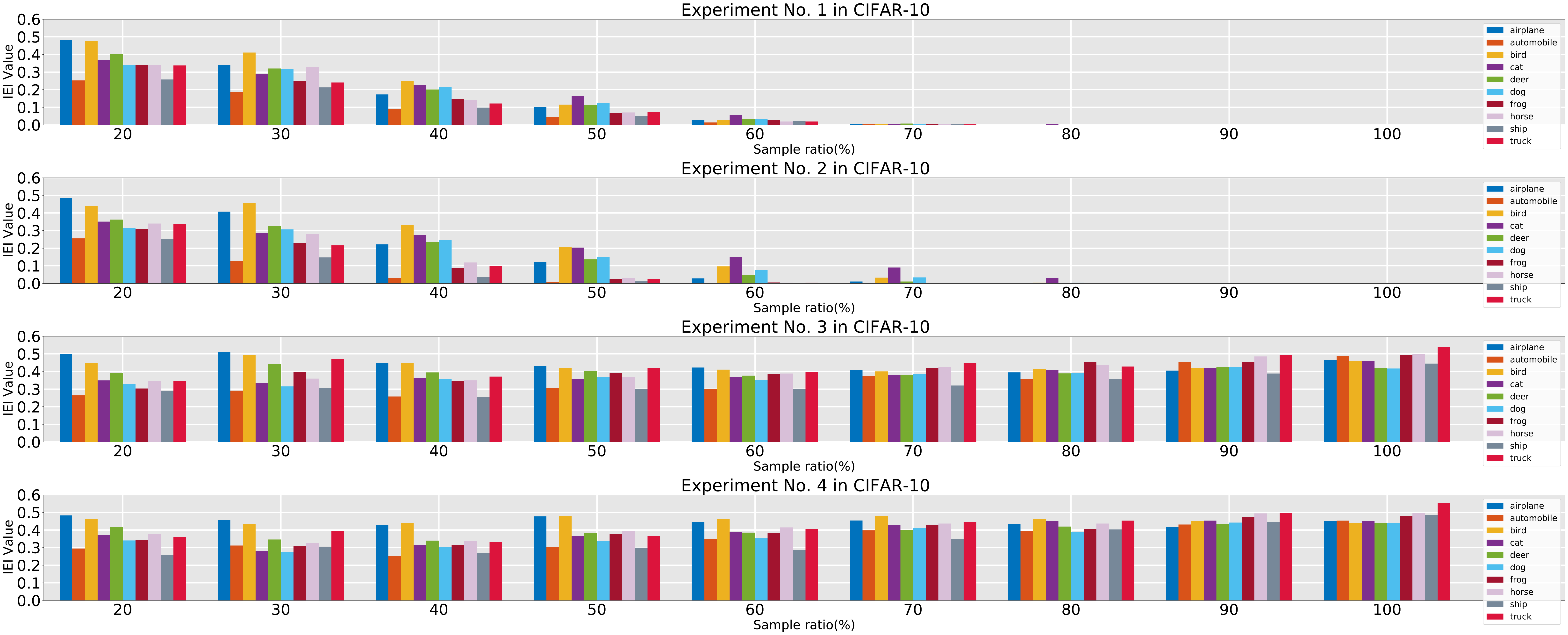}
		\caption{}
		\label{Figure4e}
	\end{subfigure}
	\centering
	\begin{subfigure}{0.96\linewidth}
		\centering
		\includegraphics[width=0.95\linewidth]{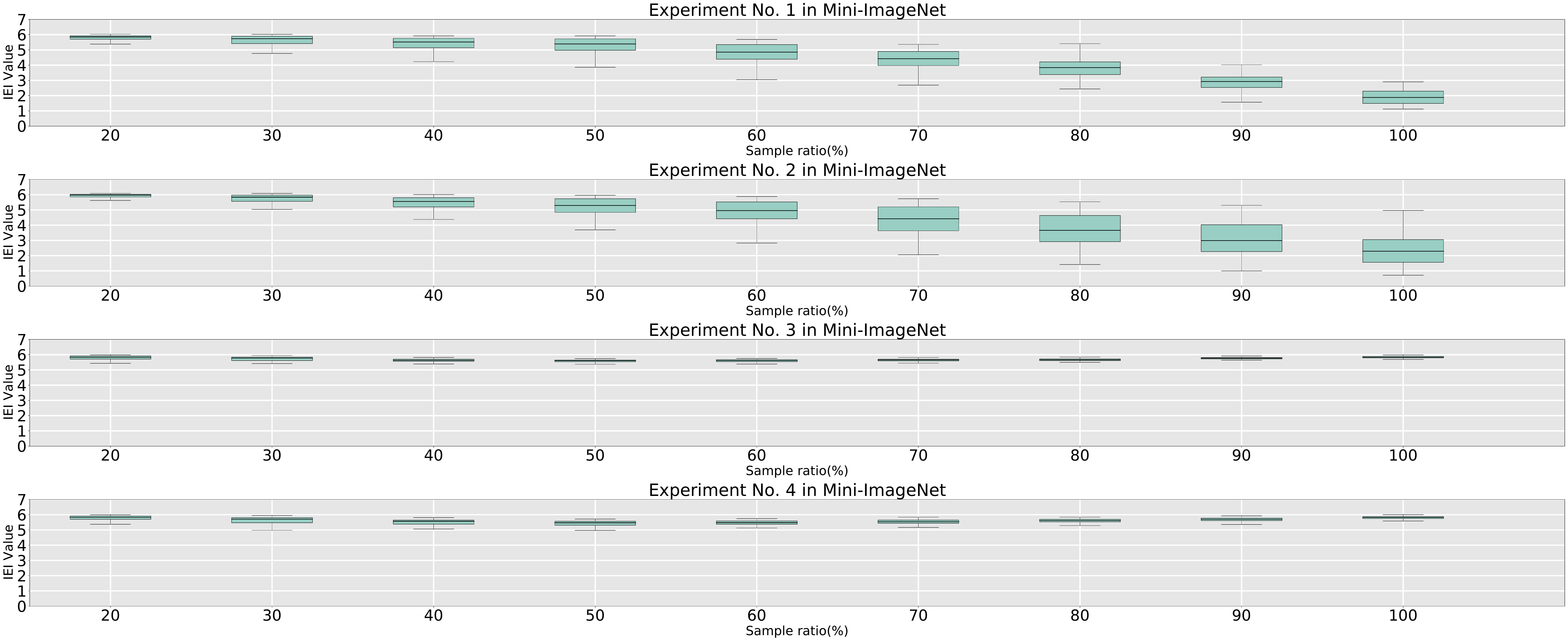}
		\caption{}
		\label{Figure4f}
	\end{subfigure}
	\centering
	\caption{Discussion and Analysis on the amount of information. (a) In an addition experiment, the distribution of IEIs of various samples in CIFAR-10 poolset, and the mean and variance of each class are different. (b) Show the distribution of various IEIs in Mini-ImageNet, which is similar to that in (a). (c) and (d) show the results of the control experiment in CIFAR-10 and Mini-ImageNet, respectively. It can be seen that the scheme without budget restriction can add the samples more efficiently. (e) and (f) further show the impact of different budget schemes on the amount of information.}
	\label{IID_explain}
\end{figure}

To solve this problem, we have made a more analysis base on the previous experiment and found that in addition to the differences in performance, the value of the IEI showes an interesting pattern for the distribution of various samples to be selected, as shown in the Fig. \ref{Figure4a} and Fig. \ref{Figure4b}.

This phenomenon shows that different classes of samples in pool set have a different amount of information distribution, which means the samples’ amount of information to be selected in some classes is generally high, while others are generally low; significant differences exist in the internal amount of information among the samples in the pool set with some classes, while others are not.

Based on the above analysis, we propose the "two-tier amount of information" theory for data-driven algorithm to construct the relationship between amount of information and practice.

\begin{enumerate}
\item The relative relationship between different classes determines the information basis of each class, which we name the Amount of Class Information (ACI).
\item The relative relationship between the base and samples in the pool set determines their value of information, which we name the Amount of Sample Information (ASI).
\end{enumerate}

The description object of ACI is class, and that of ASI is sample. ACI reveals the information basis formed by the relative relationship between classes, which is a horizontal relative amount of information. ASI indicates the contribution of a sample to the base, which is a vertical relative amount of information. The two kinds of amount of information do not convert each other.

Therefore, the two kinds of information will play different guiding roles in practice.
\begin{enumerate}
\item The difference in the ACI determines the budget allocation scheme of various samples. The class with high ACI only needs fewer samples to achieve a good situation due to the high information density,.Because of the lack of amount of information, the class with low ACI needs more samples to realize the situation of abundant information.

\item The difference in ASI determines the scheme of samples selection. The samples with larger ASI often have more unique semantic characteristics compared with the base, and adding them to the train set can achieve better results. The samples with smaller ASI have similar semantics to the base. Adding such samples will lead to redundancy and inefficient supplement.
\end{enumerate}
It is worth mentioning that the two kinds of amount of information often show an inverse relationship, the samples from a low ACI class are generally given a higher ASI due to the low information density of base. For the classes with high ACI, a small number of samples can build a base with abundant information, and the ASI is generally low.

The above discussion constructs the relationship between amount of information and practice, in which the impact of ASI has been verified in the experiment in section \ref{sec2}, under the same budget or even less, the performance of HID dataset composed by good samples is higher than that of LID dataset. However, the sample supplement scheme with a uniform budget according to the convention does not consider the ACI, so experiments to discuss the impact of ACI should be designed.

Based on the addition experiment, the information density of the addition samples are controlled, and the comparative experiment is designed for the budget proportion of classes to form four groups of addition schemes. The specific design is shown in Table \ref{tab}. Experiment No. 1 and Experiment No. 2 can form a comparison of the influence of class proportion when adding goodset, and Experiment No.3 and No.4 can form a comparison of the influence of class proportion when adding badset. The above experiments are carried out on CIFAR-10 and Mini-ImageNet.

\begin{table}
    \centering
    \fontsize{8}{6}\selectfont    
    \caption{Comparative experiment.}
\begin{tabular}{l|cc}
    \toprule
\diagbox [width=10em,trim=l] {Addition\\ Dataset}{Budget \\Scheme} & Unbalance & Balance \\
\hline
\\
Goodset & Experiment No. 1 & Experiment No. 2   \\
\\
Badset & Experiment No. 3 & Experiment No. 4 \\
    \bottomrule
\end{tabular}\vspace{0cm}
    \label{tab}
\end{table}

The results of the experiments are shown in Fig. \ref{Figure4c} and Fig. \ref{Figure4d} Specifically, Experiment No. 1 achieves the most efficient addition, Experiment No. 2 takes a little second place, Experiment No. 4 has a relatively low addition efficiency, but Experiment No.3 is even worse. This shows that selection with an evenly class strategy can not provide the highest efficiency when selecting goodset or badset, which reflects the impact of ACI on sample selection strategy.

Moreover, the IEIs distribution of pool set can be interpreted as the demand of the current algorithm for each class of samples. Fig. \ref{Figure4e} shows the visualization results of the distribution changes during the addition process of CIFAR-10. It can be seen that the IEI distribution of Experiment No.1 tends to have a low value and average state when the budget reaches 70\% due to the efficient and adaptive addition, while in Experiment No.2, due to the even selection, some high ACI classes reach the low index state earlier, while the low ACI class always maintains the high Indicator state. In Experiments No.3 and No.4, due to the addition of badset, the Indicator of samples in pool set has always been relatively high. Fig. \ref{Figure4f} shows the visual results of the supplement process of Mini-ImageNet. Since there are 100 classes, the box diagram is used to show the distribution of each class's indicators in each round of addition. It can be observed that the overall trend is similar to Fig. \ref{Figure4e}.

\subsection*{Bias-fitting: A noticeable cause of OOD condition}
The OOD experiment verifies the existence of the cross-domain, which confirms that using a small number of positive migration samples to participate in training can make the network have better generalization performance, thereby reducing the degree of OOD. However, when a network misclassified an OOD class, we can't know the main cause. This section will also use these two datasets mentioned above, NICO-Animal and CSE, to explore the cause of OOD.

\begin{figure}[htbp]
	\centering
	\begin{subfigure}{0.85\linewidth}
		\centering
		\fbox{
		\includegraphics[width=0.425\linewidth]{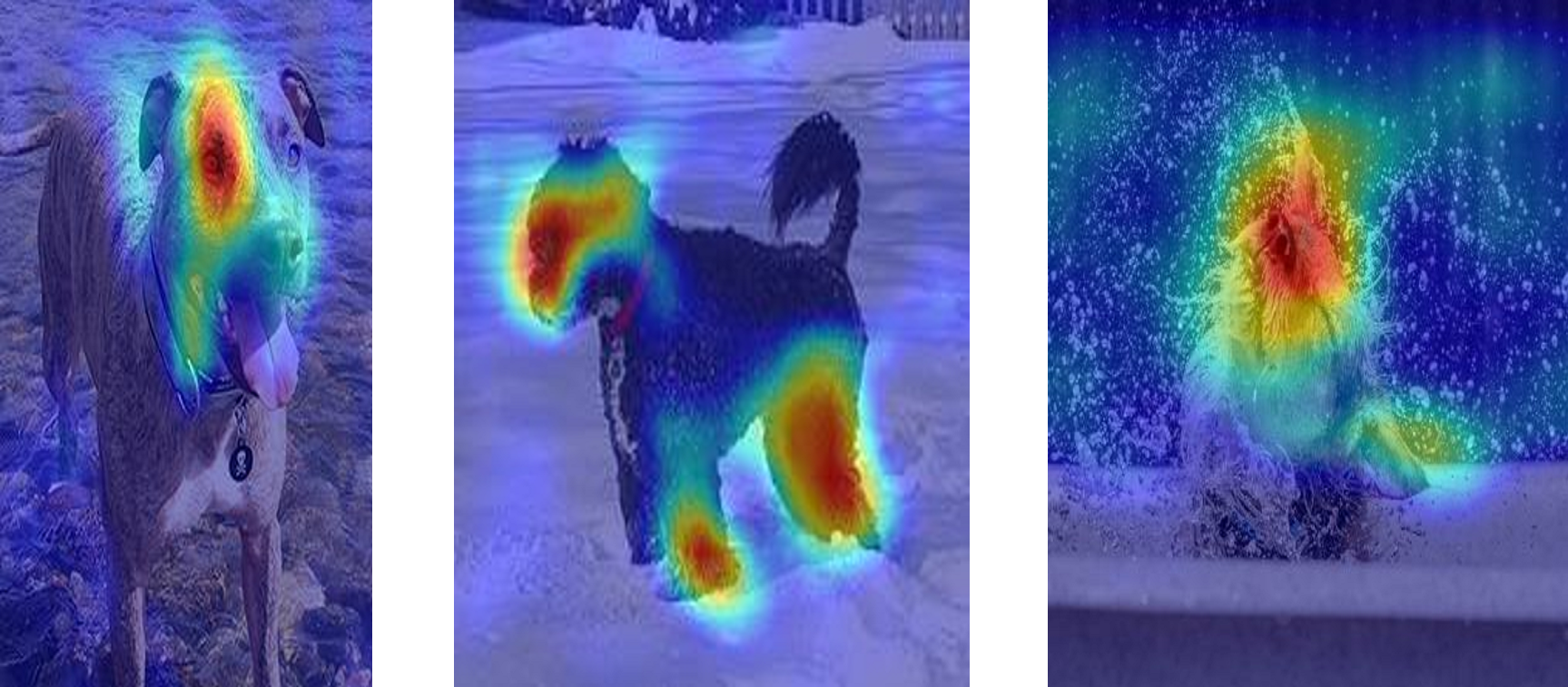}
		\includegraphics[width=0.425\linewidth]{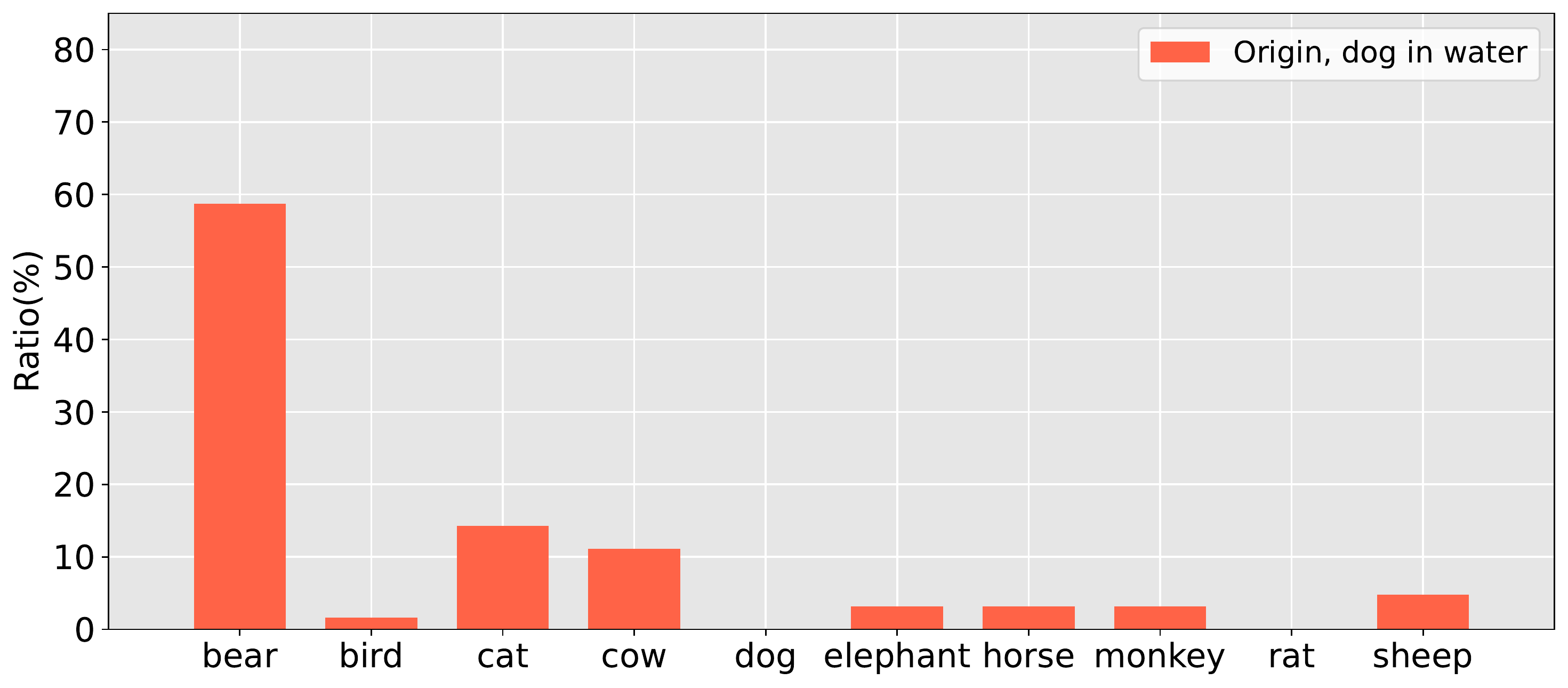}
		}
		\caption{}
		\label{Figure6a}
	\end{subfigure}
	\centering
	
	\begin{subfigure}{0.85\linewidth}
		\centering
		\fbox{
		\includegraphics[width=0.425\linewidth]{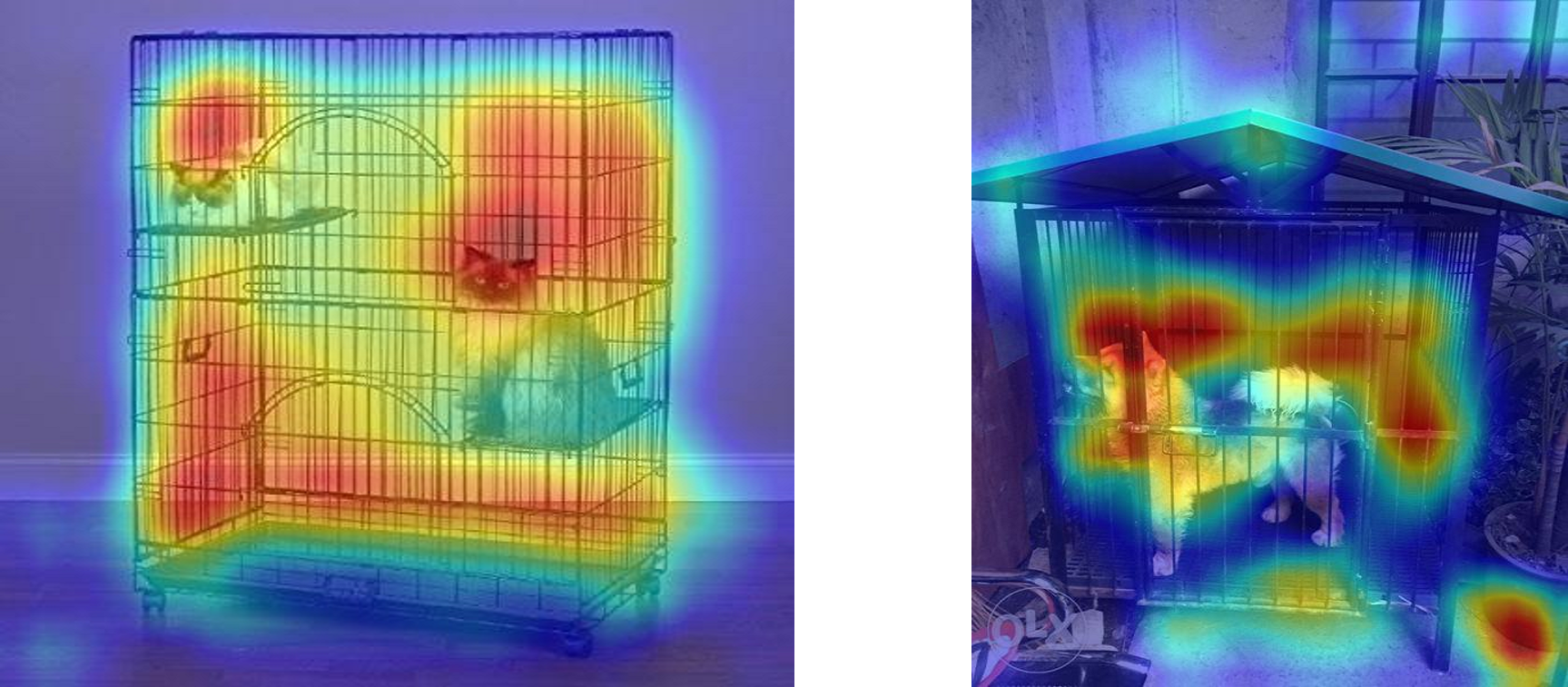}
		\includegraphics[width=0.425\linewidth]{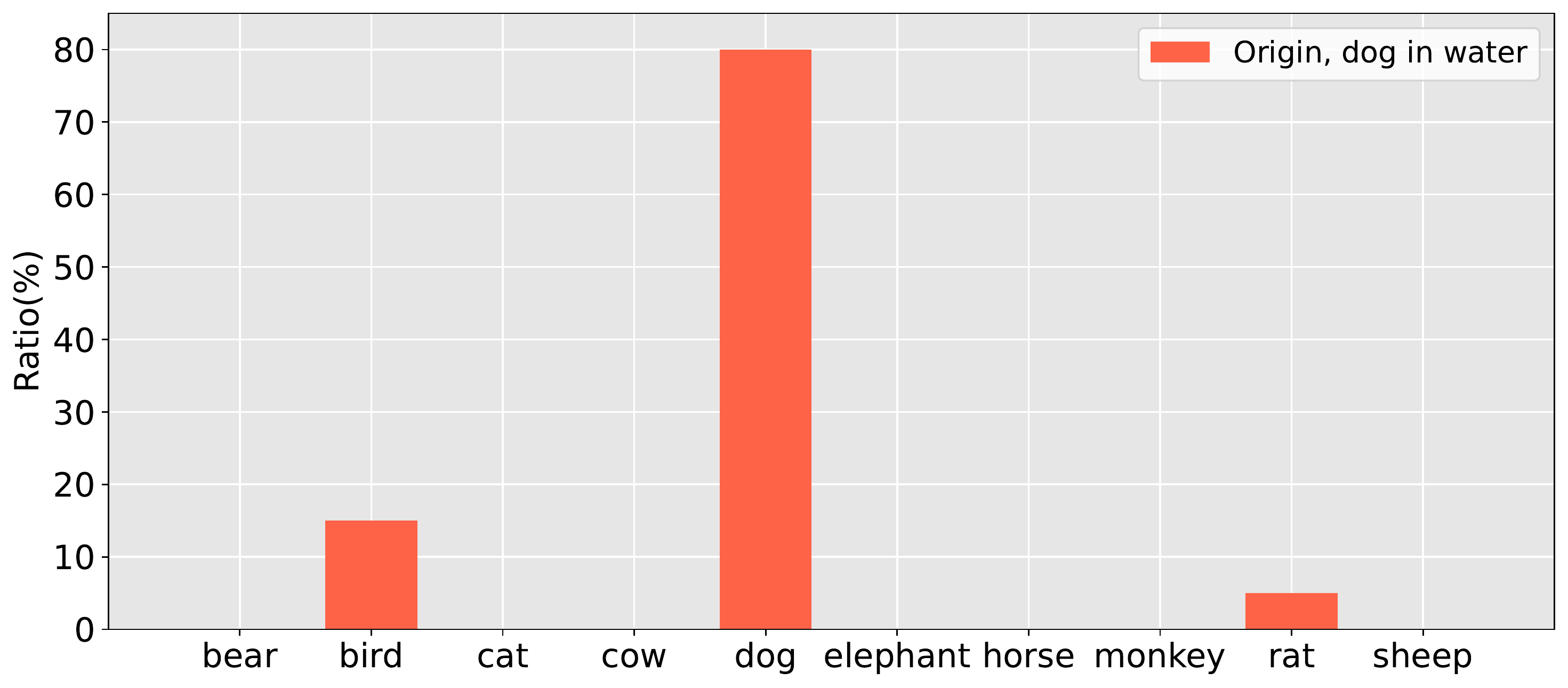}
		}
		\caption{}
		\label{Figure6b}
	\end{subfigure}
	\centering

	\begin{subfigure}{0.43\linewidth}
		\flushright
		\includegraphics[width=0.9\linewidth]{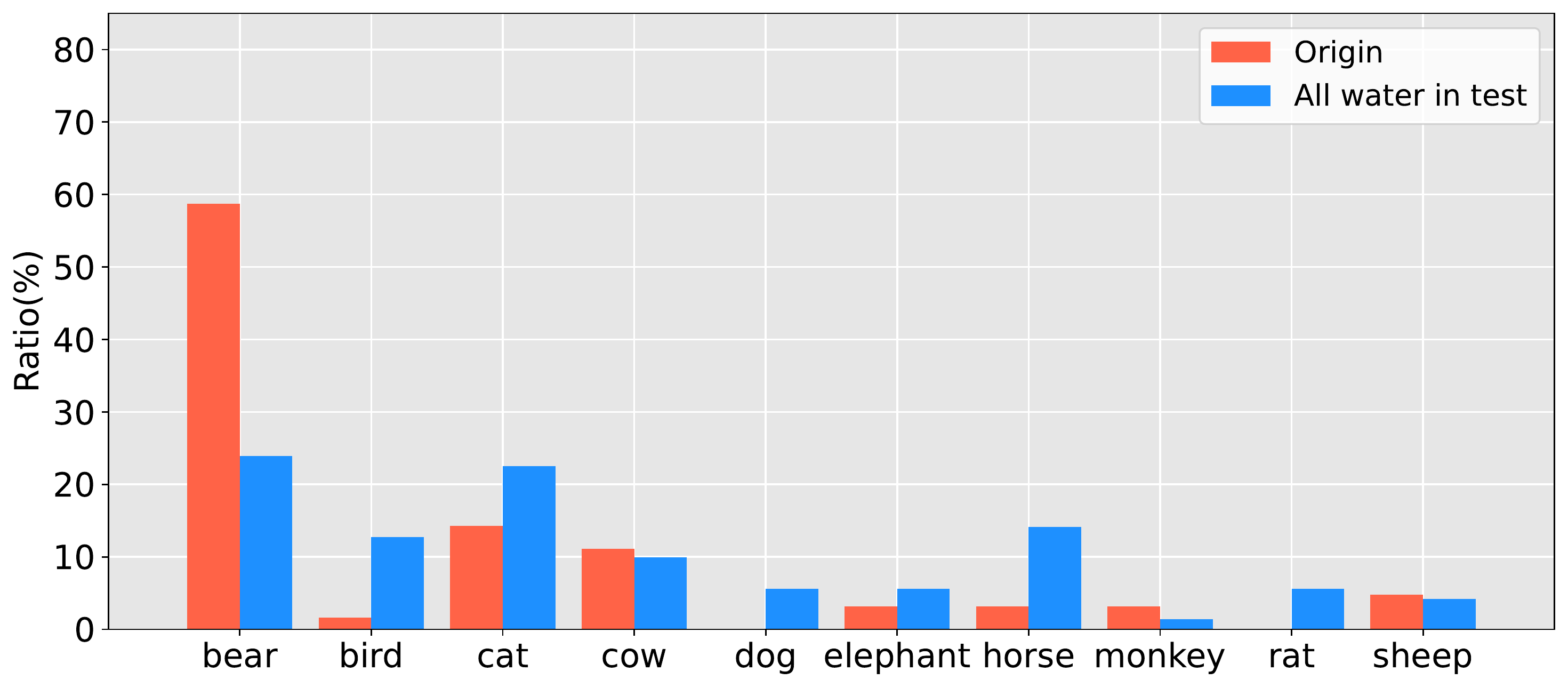}
		\caption{}
		\label{Figure6c}
	\end{subfigure}
	\centering
	\begin{subfigure}{0.43\linewidth}
		\flushleft
		\includegraphics[width=0.9\linewidth]{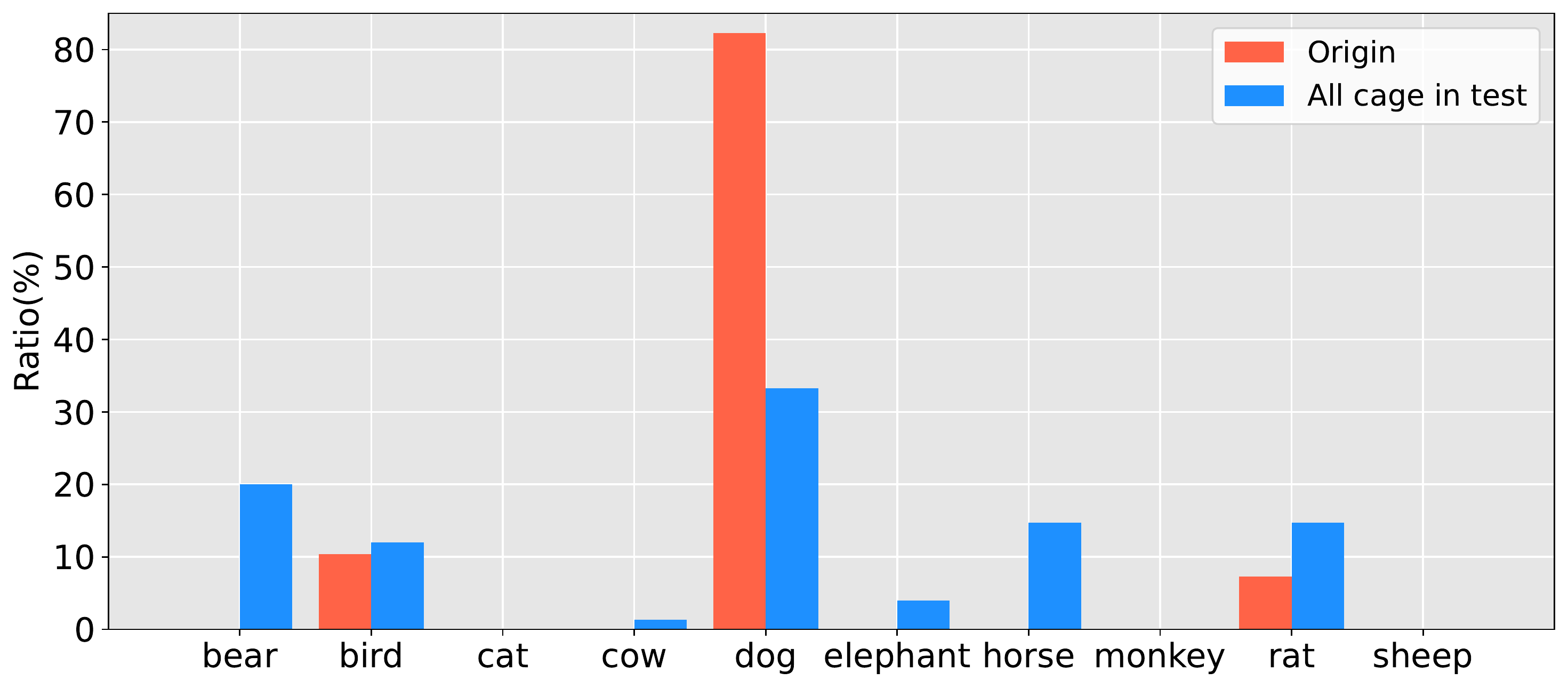}
		\caption{}
		\label{Figure6d}
	\end{subfigure}
	\centering

	\begin{subfigure}{0.85\linewidth}
		\centering
		\fbox{
		\includegraphics[width=0.425\linewidth]{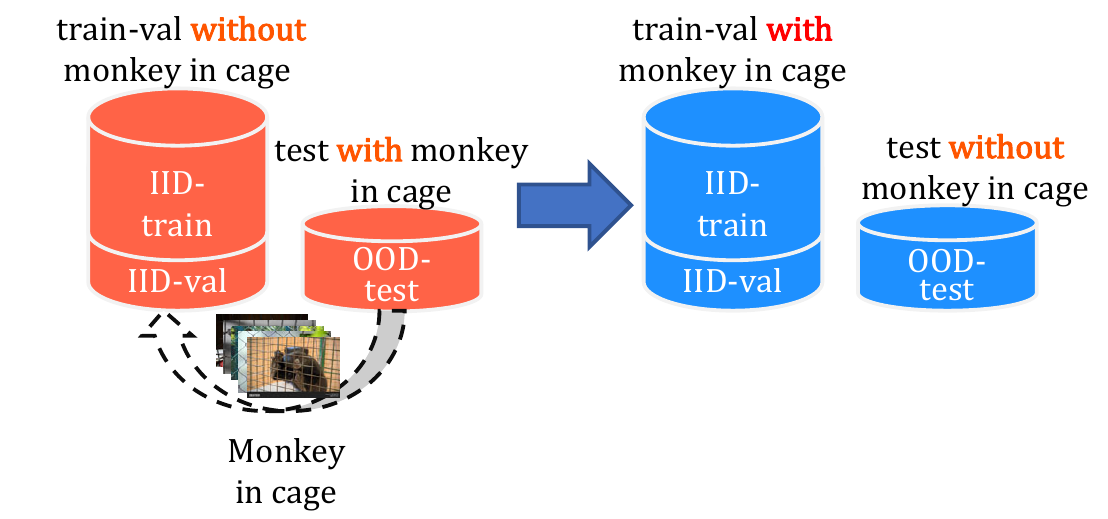}
		\includegraphics[width=0.425\linewidth]{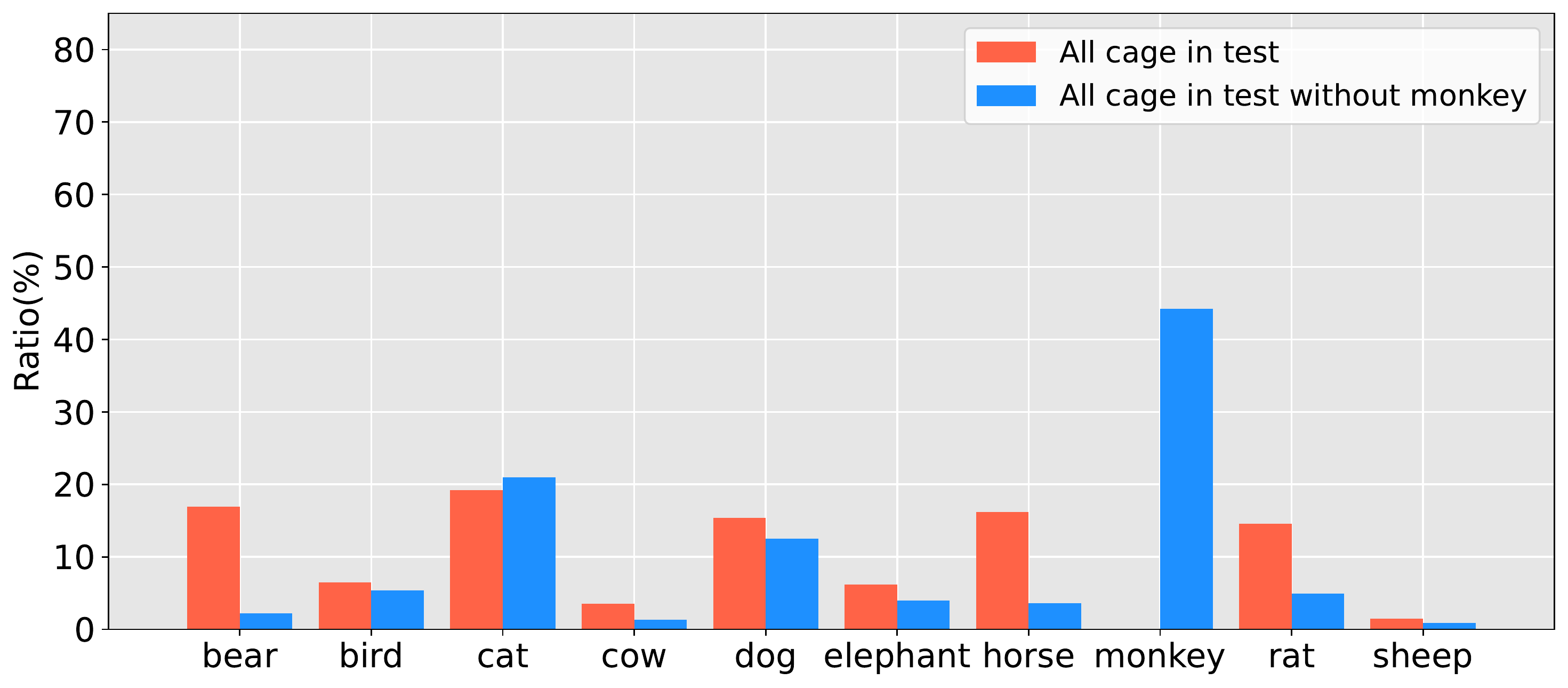}
		}
		\caption{}
		\label{Figure6e}
	\end{subfigure}
	\centering
	\begin{subfigure}{0.9\linewidth}
		\centering
		\flushleft
		\includegraphics[width=0.9\linewidth]{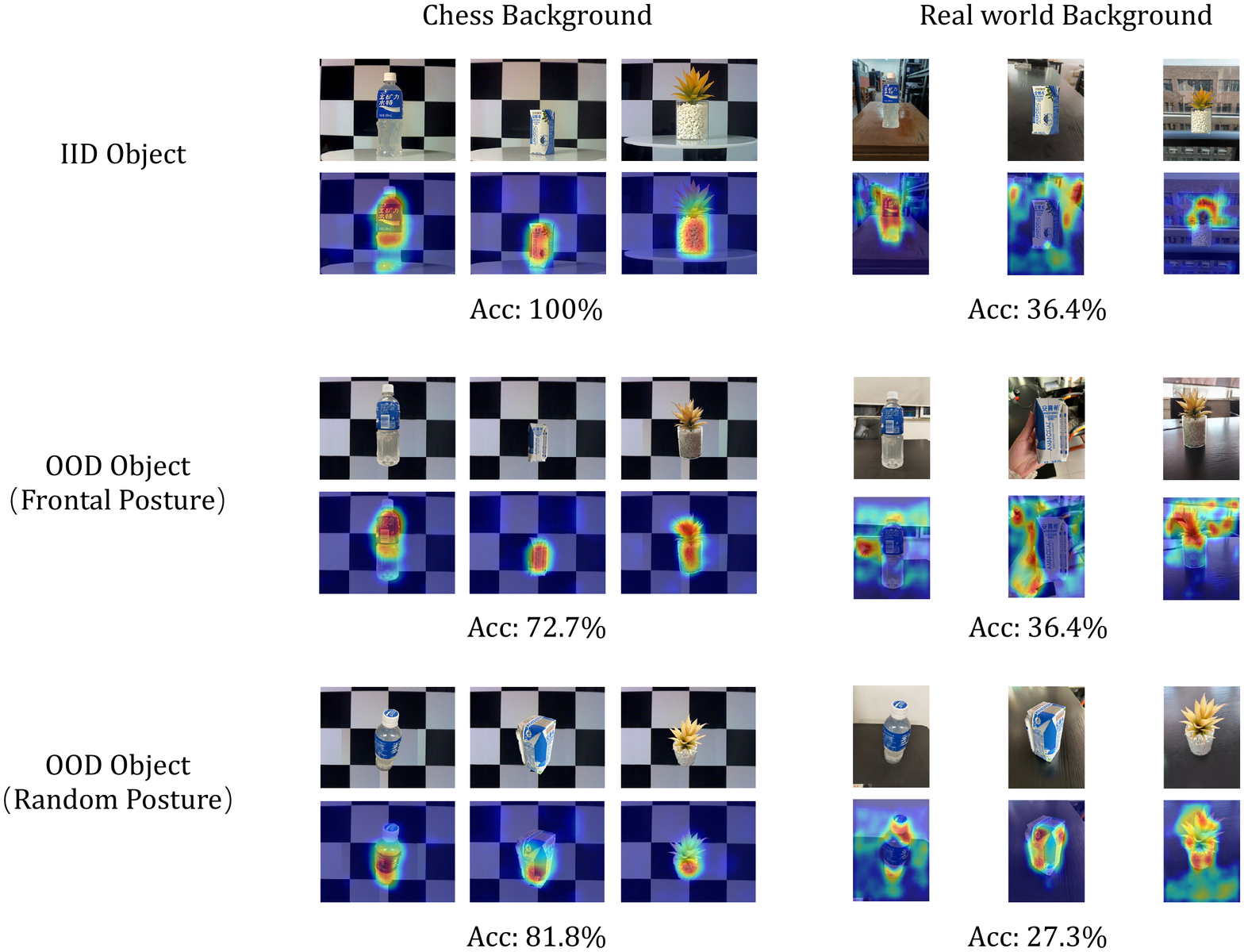}
		\caption{}
		\label{Figure6f}
	\end{subfigure}
	\centering
	\caption{OOD is mainly caused by bias-fitting. (a)The heatmaps and error distribution of "dog in water" class. (b)The heatmaps of "cat in cage" and "cat in cage" samples, which all focused on cage. The right figure shows error distribution of "dog in water" class. (c)After decoupling "in water" factor, the error distribution becomes uniform. (d)After decoupling "in cage" factor, the error distribution becomes uniform. (e)Let "monkey in cage" participate in training, the error distribution becomes "biased". (f)When the background is chess, whatever the object is from IID or OOD sample, the accuracy rises; when using real world background, the accuracy falls.}
	\label{OOD_explain}
\end{figure}

\noindent\textbf{Analysis of OOD based on NICO-Animal dataset.}
The samples in NICO-Animal dataset are all real-world , but the foreground (object) or background is different between each class, so every class is OOD with each other. We take the network trained in the OOD result part\ref{OOD_result} to perform the following exploration. After testing on the test set, we find that most classes with high error rate have unbalanced error distribution. Then, we use Grad-CAM\cite{8237336} as a visualization tool, which can be used to visualize the network decisions as a heatmap. Here are the following appearances:

\begin{enumerate}
\item Most of the misclassified classes can correctly focus on the object of the sample, and many of them are affected by the background and thus appear "biased", which is named as bias-fitting. In detail, taking context "dog in water" as an example, when looking at the heatmap of the misclassified samples from the test dataset, it is found that the network successfully notices the dog, as shown in Fig. \ref{Figure6a}, but divides them into other classes, and the distribution is shown in Fig. \ref{Figure6a}. Subsequent investigations finds that the three classes with the highest error rate all has "in water" or "in river" factor in the train set and validation set and participated in training.
\item The bias-fitting phenomenon is more serious in the "in cage" factor in the test set, even paying attention to the occluders (cages) and hardly paying attention to the real objects. The specific performance is as follows: Check the heatmap of the misclassified test set samples, as shown in Fig. \ref{Figure6b}, it is found that the "dog in cage" context is in the train set and validation set and participates in the training, and after checking the heatmaps of samples in "dog in cage", it is found that the cages are also noticed, as shown in the right heatmap in Fig. \ref{Figure6b}. It also has a "biased" distribution, also as shown in Fig. \ref{Figure6b}.
\end{enumerate}

Based on the above appearances, we propose a hypothesis that may lead to the OOD appearances in the NICO-Animal dataset: Bias fitting is the main reason that causes the network to "bias" the classification of test samples. It includes the coupling information of object and background, which we call element coupling, and abnormal occasions such as occlusion. Among them, occlusion will cause more serious bias-fitting, so that the network can't pay attention to the object, and even make decisions totally on these occluders.

Are these bias-fitting appearances, especially error distribution "biases", really caused by the samples that appear in the train? To answer this question, we design several experiments based on the idea that if the biased samples are removed from the train, these "biases" will disappear, thus proving that both the bias-fitting and the "biases" are caused by them.

\begin{enumerate}
\item To verify the bias-fitting caused by the factor coupling, we design an "in-water element decoupling" experiment. We construct a new set of datasets based on the train, validation, and test: exchange all samples with "in water" element in the train and validation with other element in test which in the same class and with similar amounts, then train and test like the beginning of this section.
\item To verify that some occasions of the "in cage" element lead to bias-fitting, we design an "in-cage element decoupling" experiment. We exchange all samples with "in cage" element in the train with train and validation like we do on "in-water element decoupling", constructing train and validation without "in cage", then train and test like the beginning of this section. 
\end{enumerate}

The result of we design a "in-water element decoupling" is shown in Fig. \ref{Figure6c}, the error distribution of "dog in water" is becoming uniform. It shows that there exists bias-fitting and leads to misclassification, causing unbalanced error distribution.

The result of we design a "in-cage element decoupling" is shown in Fig. \ref{Figure6d}, it is found that the error distribution becomes uniform. Therefore, we exchange "monkey in cage" into train and validation, then repeat the experiments. It shows that the errors in the second group are all shifted to"monkey in cage", as shown in Fig. \ref{Figure6d}, and even "dog/bird/rat in cage" which are initially in the train and validation also become increasingly misclassified as monkeys, because of "biases" caused by "monkey in cage", as shown in Fig. \ref{Figure6e}, indicating that "monkey in cage" greatly affects the classification. These experiments show that it is indeed the "in cage" occlusion that causes bias-fitting, and will reduce generalization to this class.

\noindent\textbf{Analysis of OOD based on CSE dataset.}
Compared with NICO-Aniaml, the CSE dataset has an obvious feature: background element and angle element of train set are single, but diverse in test set. It is easy to put forward a hypothesis: background or angle may affect its generalization performance, which is also bias-fitting. To test this, we edit some test samples to four patterns: IID background IID angle, IID background OOD angle, OOD background IID angle, and OOD background OOD angle. Split the datasets, take the network trained in OOD result part\ref{OOD_result} then test on these samples, the result is shown in Fig. \ref{Figure6f}.

We can see that the angle has a weak impact on test accuracy, but when the background of the OOD sample is changed from the original real scene to the IID background, the accuracy is greatly improved, and heatmaps show the network focuses mainly on foreground, and merely a bit on the background. We also design a comparative experiment, replacing the IID sample background with the OOD background. We find that the accuracy is significantly reduced, and the heatmaps become sparse, as shown in Fig. \ref{Figure6f}. The above results prove that the network trained by the IID datasets with a single background will cause bias-fitting, and some uncertain factors in the complex background in the test can greatly affect the classification of the mode, which causes OOD.

\subsection*{Summary}
After discussion and experimental verification, we expand the problem of the amount of information in the field of IID into the theory of "two-tier amount of information", and put the practical guidance on the relationship between various classes in the dataset and the relationship between supplementary samples and base. The causes of cross-domain problems in OOD are summarized as bias-fitting, and the focus of practical guidance is placed on judging whether the elements in the cross-domain sample are irrelevant.

\section{Conclusions}\label{sec4}

The work of this paper focuses on “do neural networks always perform better when eating more data?”. The experiments are carried out from the amount of information derived from sample and cross-domain sample similarity. The experiments prove that, in the case of IID and OOD, the network is not always performance better when eating more data. In the IID addition experiment, the performance of the network 30\% HID and 73\% LID on the CIFAR-10 dataset is the same, but the difference in the number of data is about 21500. On the OOD cross-domain experiment, the training results of 40\% positive migration data can be better than the results of 60\% negative migration data on both CSE and NICO-Animal datasets. 

What's more, we analyze the performance improvement of data-driven algorithms from the perspective of data, and provide a new idea for data-centric research. This idea has been effectively verified in image classification tasks. We believe that this idea has strong universality and application prospects, and is worthy of further expansion to multi-task and multi-service. In the future, based on the work of this paper, we will continue to carry out sample theoretical research in response to the current industry demand for sample analysis.

\section{Methods}\label{sec5}
\textbf{Datasets.}\label{Datasets} We took the two public datasets of CIFAR-10 and Mini-ImageNet in the IID experiment section. The train set and test set of the two datasets presents IID, which is in line with our experimental needs. CIFAR-10 is a small dataset for identifying a general object, which is widely used for the task of image classification. CIFAR-10 contains 10 categories of RGB color graphics: airplane, car, bird, cat, deer, dog, frog, horse, boat and truck. The picture size is 32$\times$32, and there is a total of 50000 train pictures and 10,000 test pictures in the data. 

The Mini-ImageNet dataset is the Google DeepMind team Oriol Vinyals and others extracted based on ImageNet. The Mini-ImageNet contains a total of 60000 color images in total, with 600 samples per category. In the experiments, the train set and test set of this dataset are divided into 5: 1. 

We have adopted CSE and NICO-Animal two datasets in the OOD experiment. CSE is a full elements dataset that we collect to explore the impact of sample elements on sample informativeness, which can support data-centric work. It rotates and shifts the samples through a six-degree-of-freedom platform, with chess images as the background. There are five objects in each category, one image is collected every 5 degrees, one group is collected every 10 cm, and three groups are collected. We take a total of 11 categories of targets, each with 1080 (5$\times$72$\times$3) images in JPEG format. To meet the cross-domain attributes, in the test set, we place objects on real scenes such as tables and chairs, take images with smartphones, and randomly take images at different angles, totaling 100 images of each category. In the experiment, we don’t do anything with CSE because it meets the OOD requirements.

NICO is a dataset especially collected for image classification in OOD scenarios, which violates the assumption that most machine learning requires IID for data, and can support much research like transfer learning and domain adaptation. There are two superclasses in NICO, Animal and Vehicle, in which we use the Animal superclass. The Animal superclass contains ten different categories, each containing 9-10 contexts. Contexts represent the posture, motion, or background information of the sample. There are 83-215 samples per context, and the average number of each category is about 1300. Different contexts are different data distributions, so we think each context is OOD to the other. Based on the number of images in the specific context, we divide it into 7:3 scales, i.e., there are 7 contexts in each train dataset, 3 contexts in the test dataset, and the ratio of the number of train-test is about 7:3. In this way, the NICO-Animal dataset is the one we used in our experiments to meet the OOD needs.

\textbf{Experiment settings.}\label{Experiment settings} We have done a lot of experiments to prove our conjecture, and we are equipped with reliable physical and software resources to ensure the validity and scientificity of the results. Specifically, in terms of physical experimental conditions, we use four NVIDIA RTX 3080Ti GPUs. In terms of software experimental conditions, we use Ubuntu 18.04 system, equipped with CUDA 11.3, GCC 7.3, Python 3.7, Pytorch 1.10.2 and OpenCV 3.4.2.

In the IID experiment, ResNet-18 is used as the backbone network. On CIFAR-10, batchsize is set to 256, 80 epochs are set for classification network training. In each epoch, the initial learning rate is set to 0.01, and the learning rate decreased ten times at the 30th and 60th epochs. Momentum and weight decay are set to 0.9 and 0.0005, respectively. For Mini-ImageNet, the batchsize is set to 128, 120 epochs are set for classification network training, with an initial learning rate of 0.1 in each epoch and a ten-fold reduction in the learning rate at the 30th, 60th and 80th epochs. The momentum and weight decay settings are the same as above. In the case of the increase sample experiment, we first initialize the base class dataset using 5000 randomly selected images. For each samples selection cycle, we select 5000 images from the pool of samples to be selected and place them in the train set until the train set increased to 50000. We use a similar method in the reduction experiment.

In the OOD experiment, to prove the generalization of the method, we used three different backbones, VGG-16\cite{simonyan2014very}, ResNet-18\cite{he2016deep}, and WRN-22-8\cite{zagoruyko2016wide}. They are all very classical network frameworks in deep learning, representing three different sizes of networks: small, medium, and large. ResNet-18 is currently the most popular image classification network, so we use it as the basic feature extractor, and the two splits are validated on VGG-16 and WRN-22-8 respectively. Due to the large differences in the size of the three networks (i.e. the parameters), the selection of the hyperparameters during training is also different. In VGG-16, batch size is 150 per epoch, learning rate in CSE is 0.0001, and the learning rate in NICO-Aniaml is 0.01. In ResNet-18, 50 batch sizes per epoch, 0.001 in CSE and 0.01 in NICO-Aniaml. In WRN-22-8, 25 batch sizes per epoch, 0.01 for CSE and NICO-Aniaml, and 100 epochs for all three backbones. Images are uniformly resized to 224$\times$224 before they are entered into the network. The network optimizer is SGD and the momentum is 0.9. For a better fit of the network, the learning rate of every 30 epochs decreases by 0.1 times.

\textbf{IEIs.} In the IID experiment, we selecte Distance-Entropy, Probability-Entropy, and Metric as IEIs.
In the feature distribution of the dataset, similar samples have high similarity and will be relatively clustered in the feature space. The Distance Entropy IEI consider that high-dimensional features are between various distributions and the samples at various edges have large distance entropy and are high information samples. Samples that are close to a certain class of feature prototype and have small distance entropy are low information sample. To obtain the Distance Entropy, we firstly map the train set to the feature space with the help of the feature extractor $\mathbb{E}$ trained by the train set, then obtain the high-dimensional feature vector$E_i$ of the train set. The high-dimensional feature vectors of the samples in the training subset of each class in the base are weighted and averaged to obtain the prototype\cite{snell2017prototypical} $Proto_i$ of each class:
\begin{equation}
    Proto_i=\frac{1}{n} \sum_{i=1}^{n} E_{i}
\end{equation}
Then use this feature extractor to extract the high-dimensional features of the sample set to be evaluated, calculate the Euclidean distance between each sample and its prototypes, and obtain the distance vector $L=[a_1,a_2,...,a_n]$, where n is the number of categories of samples in the dataset. Negative the distance vector and normalize it to get the prototype probability vector:
\begin{equation}
P=\left[p_{1}, p_{2}, \ldots \ldots, p_{n}\right]
\end{equation}\begin{equation}
p_{i}=\frac{e^{-a_{i}}}{\sum_{j=1}^{n} e^{-a_{j}}}
\end{equation}Using the prototype probability vector of the sample, the distance entropy of the sample is obtained:
\begin{equation}
E\left(x_{i}\right)=\sum_{i=1}^{n} P_{i} \log _{2} \frac{1}{P_{i}}
\end{equation}
In the probability distribution obtained by the output of the network, the output probability distribution of samples with large uncertainty is relatively average, while the probability of samples with small uncertainty belonging to a certain class is particularly high. The Entropy information evaluation index considers that the output probability distribution is relatively uniform, i.e., the samples with large entropy are high information samples, and the probability of outputting a certain class is particularly large, i.e., the samples with small entropy are low information samples. With the help of the network trained by the train set, the samples to be screened are tested, and the output $n$-dimensional vector $Output=[o_1, o_2,..., o_n]$ is obtained, and the output Output is normalized by the $Softmax$ function to obtain the sample belonging to various classes. The probability distribution of :
\begin{equation}
P=\left[p_{1}, p_{2}, \ldots \ldots, p_{n}\right]
\end{equation}
The probability entropy of the sample is calculated as:
\begin{equation}
E\left(x_{i}\right)=\sum_{i=1}^{n} P_{i} \log _{2} \frac{1}{P_{i}}
\end{equation}
In the feature distribution of the dataset, similar samples have high similarity and will be relatively clustered in the feature space. The Metric IEI considers that the high-dimensional features of a certain class are far away from the prototype of this class of feature, i.e., the samples with far feature distance are high information samples, which are close to a certain class of feature prototype, i.e., samples with a short feature distance are low information samples. When obtaining the distance entropy, first, a feature extractor $\mathbb{E}$ is trained on ResNet-18 from the train set, and the feature extractor is used to map the train set to the feature space to obtain the high-dimensional feature vector $E_i$ of the train set. The high-dimensional feature vectors of the samples in the subset of each class in the train set are weighted and averaged to obtain the prototype $Proto_i$ of each class:
\begin{equation}
Proto_i=\frac{1}{n} \sum_{i=1}^{n} E_{i}
\end{equation}
Then use this feature extractor to extract the high-dimensional features of the pool set to be evaluated, calculate the Euclidean distance between each sample and the prototype of the class to which it belongs, and obtain the distance $L$ of the sample.

\textbf{OOD Theory.}\label{OOD Theory} In the OOD method, we first use the train domain and test domain of the dataset to train a feature extractor $\mathbb{E}$ on ResNet-18. $\mathbb{E}$ is equivalent to a mapping function, which maps the sample (x, y) to the high-dimensional feature space and becomes the feature vector $f_i = \mathbb{E}(x_i)$. After extracting all the features of the test domain according to the category, the prototype of each category of feature is obtained.
\begin{equation}
    p = \frac{1}{N}\sum(f_i)
\end{equation}
 Euclidean distance is used to measure the similarity between the feature vector $f_i$ of each category of samples in the train domain and the feature prototype $p_c$ of this category,
\begin{equation}
    d_{eu}=\sqrt{\sum\limits_{i=1}^{N}(f_i^c-p^c)^2 }
\end{equation}
Arrange the similarity from small to large, and take the first 60$\%$ negative migration samples and the remaining 40$\%$ positive migration samples as two split sets. We believe that in the dataset distributed cross-domain, the training performance of the network on the positive migration samples with similar test domains is better than that on the negative migration samples. In other words, if the samples input into the network have a negative effect on the network itself, even if the number of samples is large, the performance will not be improved. The two splits are trained on three backbones respectively. The results in Fig. \ref{OOD_Experiment} are sufficient to prove the effectiveness of our theory.

\textbf{Grad-CAM.} CAM (Class Activation Mapping) is a class of CNN visualization algorithms which can display the network decisions. Since we need to visualize the ResNet-18, the applicable algorithm called Grad-CAM is chosen, which first uses the back-propagated gradients to calculate the weights with the following formula:

\begin{equation}
\alpha_{k}^{c}=\frac{1}{Z} \sum_{i} \sum_{j} \frac{\partial y^{c}}{\partial A_{i j}^{k}}
\end{equation}

Where $c$ represents the category, $y^c$ is the logit corresponding to the category, A represents the feature map output by the convolution, $k$ represents the channel of the feature map, $i, j$ represents the horizontal and vertical coordinates of the feature map, and $Z$ represents the size of the feature map. This process is equivalent to finding the mean of the gradients on the feature map, which is equivalent to a global average pooling operation in CAM.

After the weights are obtained, the channels of the feature map are linearly weighted together to obtain the heatmap. The formula is as follows:

\begin{equation}
L_{\text {Grad-CAM }}^{c}=\operatorname{ReLU}\left(\sum_{k} \alpha_{k}^{c} A^{k}\right)
\end{equation}

Grad-CAM adds a ReLU operation to the fused heatmap, which only retains the regions that have a positive effect on the given class.

\section{Data availability}\label{sec5}
The  CSE dataset can be downloaded from \url{http://aimip.tju.edu.cn/CSEdataset.zip}. Source data are provided with this paper.
\section{Code availability}\label{sec6}
The code is freely available at \url{https://github.com/AIMIP-TJU/Sample_Information_Works} with MIT Licence. This web page contains the code dependencies, instructions, and some interactions between codes.

\bibliography{sn-bibliography}


\end{document}